\documentclass[runningheads]{llncs}

\usepackage{graphicx}
\usepackage{amssymb}
\usepackage{amsmath}
\usepackage[caption=false,font=normalsize,labelfont=sf,textfont=sf]{subfig}
\usepackage{stfloats}
\usepackage{url}
\usepackage{wrapfig}
\usepackage{bbding}
\usepackage{hyperref}
\usepackage[capitalize]{cleveref}
\usepackage{booktabs}
\usepackage[normalem]{ulem}
\useunder{\uline}{\ul}{}
\usepackage{multirow}
\usepackage{caption}
\usepackage{color}
\usepackage{enumitem}
\usepackage{subfig}
\usepackage[misc]{ifsym}

\begin{document}
%
% \title{Contribution Title\thanks{Supported by organization x.}}
\title{Straighter Flow Matching via a Diffusion-Based  Coupling Prior}
\author{Siyu Xing\inst{1,2} \and
Jie Cao\inst{3} \and
Huaibo Huang\inst{3} \\
 Haichao Shi\inst{1,2}
\and Xiao-Yu Zhang\inst{1,2}\textsuperscript{(\Letter)}}
\authorrunning{S. Xing et al.}
\institute{Institute of Information Engineering, Chinese Academy of Sciences, Beijing, China \and
School of Cyber Security, University of Chinese Academy of Sciences, Beijing, China \\
\email{\{xingsiyu, shihaichao, zhangxiaoyu\}@iie.ac.cn}\\
 \and
Institute of Automation, Chinese Academy of Sciences, Beijing, China\\
\email{\{jie.cao, huangbo.huang\}@cripac.ia.ac.cn}}
\maketitle              % typeset the header of the contribution
\begin{abstract}
Flow matching as a paradigm of  generative model achieves notable success across various domains. 
However, existing methods use either multi-round training or knowledge within minibatches, posing challenges in finding a favorable coupling strategy for straightening trajectories to few-step generation. 
To address this issue, we propose a novel approach, Straighter trajectories of Flow Matching (StraightFM). 
It straightens trajectories with the coupling strategy from the entire distribution level. 
More specifically, during training, StraightFM creates couplings of images and noise via one diffusion model as a coupling prior to straighten trajectories for few-step generation.
Our coupling strategy can also integrate with the existing coupling direction from real data to noise, improving image quality in few-step generation.
Experimental results on pixel space and latent space show that StraightFM yields attractive samples within 5 steps.
Moreover, our unconditional model is seamlessly compatible with training-free multimodal conditional generation, maintaining high-quality image generation in few steps.
\end{abstract}
%
% \linenumbers

\section{Introduction}
\begin{figure}
\begin{minipage}{0.475\textwidth}
    \centering
    \includegraphics[scale=0.3]{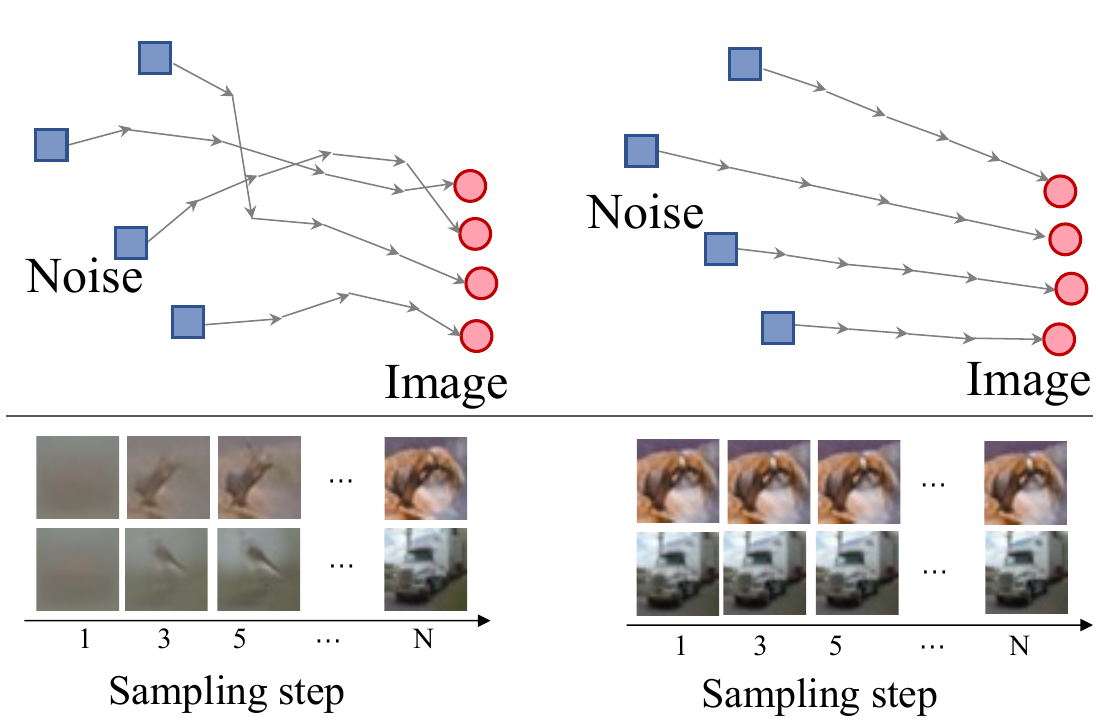}
    \vspace{-1mm}
    
		\begin{minipage}[t]{0.25\linewidth}
			\centering		
\hspace{-17.5mm}		(a)	Flow matching
		\end{minipage}%
	{
		\begin{minipage}[t]{0.25\linewidth}
			\centering
			\vspace{-2mm}(b) Ours
		\end{minipage}%
	}
    \caption{
     Illustration about vanilla flow matching and ours. StraightFM (top-right) supports straighter trajectories than vanilla flow matching (top-left), generating high-quality images with fewer steps (bottom). }
\label{fig:illu}
\end{minipage} 
\begin{minipage}{0.475\textwidth}
    \centering
    \captionof{table}{
    Overview of key properties (few-step generation, training-free conditional generation, and one-round training) of  StraightFM compared to Rectified Flow (RF)~\cite{liu2022flow}, Optimal transport conditional flow matching  (OT-CFM)\cite{tong2023improving}, and Consistency Model (CM)~\cite{song2023consistency}.
}
  \renewcommand{\arraystretch}{0.55}
   \resizebox{1\textwidth}{!}{
\begin{tabular}{@{}llll@{}}
\toprule
Methods & \begin{tabular}[c]{@{}l@{}}Few-step\\ generation\end{tabular} & \begin{tabular}[c]{@{}l@{}}Training-free \\ conditional generation\end{tabular} & \begin{tabular}[c]{@{}l@{}}One-round\\ training\end{tabular} \\ \midrule
Ours & \Checkmark & \Checkmark & \Checkmark \\
OT-CFM & \XSolidBrush & \Checkmark & \Checkmark \\
$1$-RF & \XSolidBrush & \Checkmark & \Checkmark \\
$k$-RF (k$>$2) & \Checkmark & \Checkmark & \XSolidBrush \\
CM & \Checkmark & \XSolidBrush & \Checkmark \\ \bottomrule
\end{tabular}
}
\label{tab:duicuo}
\end{minipage}
\end{figure}

Generative models have achieved remarkable success across multiple fields. 
One class of generative models builds a mapping from data distributions to the prior distribution and learns the inverse mapping with ordinary differential equation (ODE) to generate data~\cite{chen2018neural}.
Using this paradigm, diffusion models have made significant progress in vision domains~\cite{song2021score,ho2020denoising,rombach2022high}. 
Flow matching, an emerging approach, compares favorably with diffusion models by estimating a time-dependent velocity field, avoiding hyperparameter tuning and numerous denoising steps~\cite{liu2022flow,lipman2022flow,albergo2023building}.
As depicted in~\cref{fig:illu}, flow matching methods first identify a coupling strategy that binds elements in the prior distribution to those in the data distribution when calculating objectives to learn the shortest straight path between each coupling~\cite{liu2022flow,liu2023insta}. 
Nearly straight trajectories enable one-step generation via the Euler method, minimizing time discretization errors.

However, discovering tractable and scalable couplings in flow matching is challenging. 
Optimal transport plan provides optimal solution, but this is computationally prohibitive for high-dimensional data.
 Thus, existing methods address this by using either multi-round training~\cite{liu2022flow}, or leveraging underlying information within mini-batches to find couplings~\cite{pooladian2023multisample,tong2023improving}.
 The former leads to redundant models and accumulated errors, potentially degrading image quality~\cite{alemohammad2023self}. 
 The latter does not capture the complexity across the entire data distribution, hindering applications to complex tasks in the real world.

In this work, we explore a simple yet effective solution: \textit{finding scalable and tractable couplings for flow matching via a diffusion-based coupling prior}, i.e., the Probability Flow Ordinary Differential Equation (PF-ODE) of diffusion models.
We take inspiration from recent works that use pre-trained diffusion models to train a new one with improved sample quality and inference speed~\cite{salimans2022progressive,song2023consistency}.
Flow matching and PF-ODE, despite differing in training strategies, share essential principles:
they utilize time-dependent ODEs~\cite{gao2025diffusionmeetsflow} and aim to minimize Wasserstein distance or related transport costs~\cite{de2021diffusion,kwon2022score,khrulkov2023understanding}. 
 Instead of cumbersome multi-round training, it is worthwhile to leverage existing high-quality and abundant diffusion models to provide couplings for flow matching training. 
 Moreover, since well-trained diffusion models have learned the entire data distribution, obtaining non-trivial couplings with the assistance of diffusion models could skip the procedure of solving OT problems on a large scale.

Our work explores Straighter trajectories of Flow Matching using a diffusion-based coupling prior, which is short for StraightFM.
\Cref{tab:duicuo} compares StraightFM with different baselines.
Specifically, we leverage a well-trained diffusion model to synthesize pseudo-data with ODE solvers on the PF-ODE trajectories.
Each pseudo-image and its corresponding initial noise form a coupling utilized for StraightFM training. 
Similar to most flow matching frameworks, StraightFM employs a time-dependent model to match the drift term of one ODE, which defines the velocity field pointing to the shortest straight path between each coupling
 We set the intermediate sample state as a linear interpolation of each coupling along this path.
To straighten the trajectory, StraightFM with a simple unconstrained least squares objective function guides the time-dependent model at each intermediate state to match the velocity field.
This strategy can also integrate with an existing real data to noise coupling method~\cite{lee2023minimizing}, enhancing the image quality in few-step sampling.
\begin{figure}[t]
    \centering
    \includegraphics[scale=0.375]{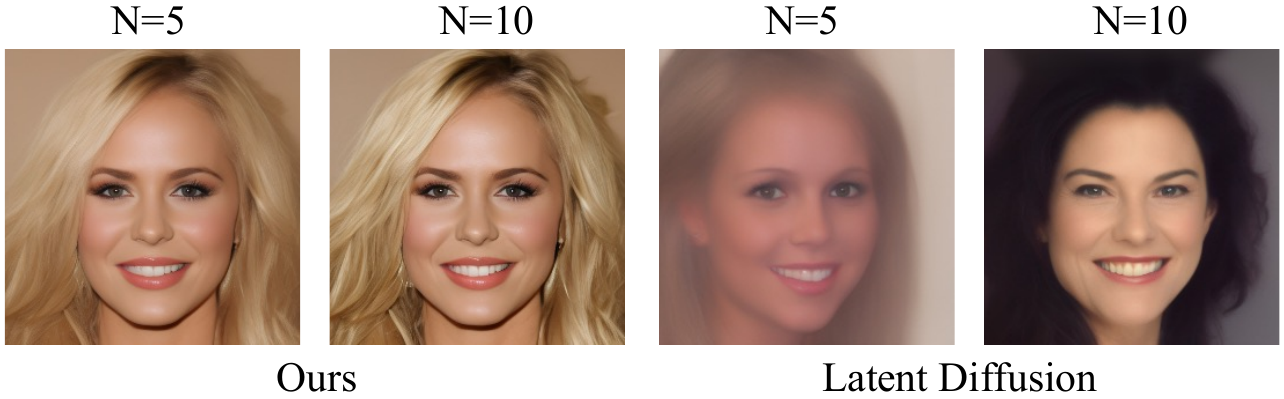}\vspace{-2mm}
    \caption{
Results on a high-resolution facial dataset. Compared to  Latent Diffusion~\cite{rombach2022high}, our model is able to generate high-quality images using 5 step. }
\label{fig:head}\end{figure}
The main contributions of this work are summarized as follows:
\begin{itemize}[leftmargin=*] 
\item We propose a simple yet effective coupling strategy for a fast-sampling flow matching with diffusion model as a coupling prior, bypassing the need for solving OT problems in minibatches or multi-round training. 
\item Extensive experiments on the pixel space (CIFAR-10 dataset) and the latent space of an autoencoder~\cite{rombach2022high} (CelebA-HQ 256$\times$256 dataset) demonstrate that StraightFM can generate high-quality images within 5 steps and even a single step. Besides, generated images of StraightFM are highly independent of sampling steps, implying straighter trajectories, as shown in \cref{fig:head}. 
\item As an unconditional fast-sampling generative model, StraightFM is flexibly adapted to diverse multimodal guided conditional generation without retraining cost, while keeping a $5\times$ speed up compared to the state-of-the-art. \end{itemize}

\section{Related Work}\label{sec:rw}
\textbf{Generative Model with ODE }
{Continuous Normalizing Flow (CNF) learns a complex data distribution by transforming a simple distribution through a series of invertible mappings specified by neural ODEs~\cite{chen2018neural}.
Training CNF can be categorized into simulation-based and simulation-free methods.
The former employs the maximum likelihood objective, leading to prohibitively expensive ODE simulations~\cite{grathwohl2018ffjord}.
 The latter, such as the PF-ODE of diffusion~\cite{song2021score,song2020ddim}, facilitates  training by specifying the target probability distribution without ODE simulations.
 Flow matching~\cite{lipman2022flow,liu2022flow,albergo2023building}, another simulation-free method, demonstrates that  CNFs can be trained effectively by matching vector fields through regression, achieving competitive performance in visual tasks.
}

\noindent
\textbf{Flow Matching}
Original flow matching uses independent data and noise couplings when calculating training objectives~\cite{lipman2022flow,liu2022flow}.
Recent works focus on the coupling strategies between data and noise samples to straighten trajectories and reduce sampling steps.
For instance, Multisample flow matching~\cite{pooladian2023multisample} and Optimal Transport conditional flow matching (OT-CFM)~\cite{tong2023improving} solve OT problems  per batch to determine couplings.
However, relying solely on minibatches limits the application to large-scale distribution.
Rectified Flow (RF)~\cite{liu2022flow}, a flow matching variant, uses multi-round training to find couplings, but recent finding~\cite{alemohammad2023self} shows that training the next generative model using synthesized data from the previous same type one iteratively leads to the quality degradation.
Unlike these methods, the coupling strategy of StraightFM leverages the mapping similarity of diffusion and FM (as seen in subsection~\ref{subsec:mapsim}), which is an inherent  property between two models, not present in GANs or VAEs~\cite{alemohammad2023self}.

\noindent
\textbf{Diffusion Model } % 
Prior works usually group the sampling of diffusion models into two categories: training-free and training-based methods.
Training free methods, i.e., discretizing stochastic differential equations~\cite{song2021score} and  PF-ODE of diffusion~\cite{song2020ddim,Karras2022edm,lu2022dpm,zhang2022deis,zhao2023unipc}, involve iterative denoising around 25 to 1000 steps.
 In contrast, training-based methods rely on additional training, such as distilling~\cite{luhman2021knowledge,salimans2022progressive,song2023consistency}, optimizing time steps~\cite{nguyen2024bellman,xue2024accelerating,zhou2023fastamed}, and combining diffusion with GANs~\cite{xiao2022tackling}, which can notably enhance the sampling speed but disrupt the
  equation structure, e.g., Consistency model~\cite{song2023consistency} is unavailable to training-free conditional generation like other diffusion models~\cite{yu2023freedom}.

With well-trained diffusion model as a coupling prior, our approach provides another direction to achieve not only few-step generation (in just one round of training)  but also training-free conditional generation, sidestepping  multi-round training and the intractability issues of OT problem in large-scale datasets.

\section{Preliminaries}\label{sec:pre}
\subsubsection{Probability Flow ODE of Diffusion Model}\hspace{0.5em} 
A notable property of diffusion models is a deterministic ODE, namely probability flow ODE (PF-ODE), whose trajectory mirrors the marginal probability densities of stochastic diffusion models. 
Considering sample $\mathbf{x}$ from $d$-dimensional Euclidean space $\mathbb R^d$ and the intermediate sample at state $t$ as $\mathbf{x}_t$, the probability density function of $\mathbf{x}_t$, given by $p_t(x):[0,1]\times \mathbb{R}^d\to\mathbb{R}^{+}$, characterizes the time-varying distribution of $\mathbf{x}_t$.
The PF-ODE is formulated as:
\begin{equation}\label{eq-pfode}
    \mathrm{d} \mathbf{x}_{t}=\left[\boldsymbol{\mu}\left(\mathbf{x}_{t}, t\right)-\frac{1}{2}g(t)^{2} \nabla \log p_{t}\left(\mathbf{x}_{t}\right)\right] \mathrm{d} t,
\end{equation}
where $\boldsymbol{\mu}(\cdot,\cdot)$ and $g(\cdot)$ are the drift and diffusion term of diffusion process, respectively. 
Typically, diffusion models employ a time-dependent model to approximate score function $ \nabla \log p_{t}\left(\mathbf{x}_{t}\right)$, and subsequently generate samples from initial noise by solving \cref{eq-pfode} with various numerical ODE solvers, such as Euler method, and Heun's 2nd order method.

\subsubsection{Flow Matching}
Suppose that noise sample $\mathbf x_0$ from a Gaussian distribution $p_0$ and $\mathbf x_1$ from a data distribution $p_1$ over $\mathbb R^d$, flow matching 
defines a time-dependent ordinary differential equation:
\begin{equation}\label{bcg-eq1}
\mathrm{d}  \mathbf x_{t}= v(\mathbf{x}_{t},t) \mathrm{d} t,t\in[0,1],
\end{equation}
where $v:[0, 1] \times \mathbb R^d \to \mathbb R^d$ represents a  vector field.
The primary goal of flow matching is to utilize a velocity model $u_{\theta}$, parameterized by $\theta$, i.e., a time-dependent neural network, to approximate $v$.
This is notably intractable for general unknown and complex distributions $p_t$.
Practically, a simplified objective of conditional flow matching~\cite{lipman2022flow,tong2023improving} effectively bypasses  $v(\cdot)$ by incorporating a latent condition $\mathbf{z}$:
\begin{equation}\label{bcg-eq3cfm}
\min_{\theta}\mathbb{E}_{t,q(\mathbf z),p_t(\mathbf x|\mathbf z)}\|u_{\theta}(\mathbf x_t;t)-v_t(\mathbf x_t|\mathbf z)\|^2,
\end{equation}
which has been proven to be equivalent to the objective of flow matching concerning the gradient of model parameters.
Usually, $\mathbf{z}$ is a random pair $(\mathbf{x}_0,\mathbf{x}_1)$, but this independent coupling strategy falls short in one-step generation~\cite{liu2022flow}.
The current coupling strategy either faces scalability challenges in solving OT problems within a batch~\cite{pooladian2023multisample,tong2023improving} or suffers from potential risks in the degradation of quality  due to multi-round training~\cite{liu2022flow}.

Our work resolves this shortcoming by a diffusion-based coupling prior to yield high-quality images with straighter trajectories and fewer steps, skipping additional straightening procedures.

\section{Straighter Flow Matching}\label{sec:method}
{ We first reveal the mapping similarity of couplings from flow matching and diffusion models to support our primary motivation in subsection~\ref{subsec:mapsim}. 
Then, we propose StraightFM in subsection~\ref{subsec:straightfm}.
Following that, subsection~\ref{subsec:expandingv} explores  an expanding version combined with real samples.
Subsection~\ref{subsec:mmgen} introduces a general training-free way to multimodal  conditional generation based on StraightFM.}

\subsection{Mapping Similarity between FM and Diffusion}\label{subsec:mapsim}

\begin{figure}[!thbp]
\begin{minipage}{1\linewidth}
        \centering
    \includegraphics[scale=0.35]{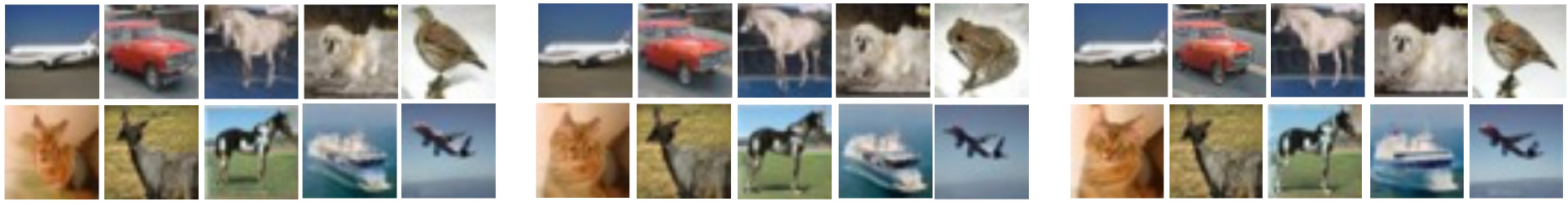}
    \end{minipage}\vspace{0mm}
    
    \centering
\begin{minipage}[t]{0.3\linewidth}
			\centering 1-Rectified Flow
		\end{minipage}
\begin{minipage}[t]{0.3\linewidth}
			\centering 2-Rectified Flow
		\end{minipage}
\begin{minipage}[t]{0.3\linewidth}
			\centering \hspace{0.5cm} EDM 
		\end{minipage}
    \caption{Samples are generated by 1-RF with RK45 (left), 2-RF with RK45 (middle) and EDM with  Heun's 2nd method (right). Corresponding images are generated from the same initial noise.}
    \label{fig:sameinput}
\end{figure}

We compare the sample difference between flow matching models (1-RF and 2-RF~\cite{liu2022flow}) and diffusion models (VP-ODE~\cite{song2021score} and EDM~\cite{Karras2022edm}) starting from the same initial noise. 
As shown in \cref{fig:sameinput} and \cref{tab:sameinput}, generated images exhibit high similarity in both numerical metrics and visual content, regardless of whether  classic diffusion model VP-ODE or state-of-the-art EDM is considered.
Although minor differences exist, most structural similarities remain unchanged.
 \begin{table}[!thbp]
\centering
{%
\caption{Comparison of 25,600 noise-image couplings from flow matching and diffusion models with identical inputs on CIFAR-10 dataset, measured by MSE and LPIPS. {Samples are generated from VP-ODE with adaptive RK45 and from EDM with 17 Heun's 2nd order steps.} $\downarrow$ means lower is better.}\label{tab:sameinput}
 \renewcommand{\arraystretch}{0.05}
\begin{tabular}{@{}lcccc@{}}
\toprule
     & \multicolumn{2}{c}{MSE$\downarrow$}              & \multicolumn{2}{c}{LPIPS$\downarrow$} \\ \cmidrule(l){2-5} 
     & VP-ODE~\cite{song2021score}   & EDM~\cite{Karras2022edm}                        & VP-ODE~\cite{song2021score}   & EDM~\cite{Karras2022edm}           \\ \midrule
1-RF~\cite{liu2022flow} & 0.0065 & \multicolumn{1}{l|}{0.0055} & 0.21        & 0.19        \\
2-RF~\cite{liu2022flow} & 0.0053 & \multicolumn{1}{l|}{0.0043} & 0.18        & 0.15        \\ \bottomrule
\end{tabular}
}
\end{table}
Recent work~\cite{gao2025diffusionmeetsflow} also highlights the equivalence between flow matching and diffusion with   theoretical insights, particularly in low-dimensional settings.
Concurrent research~\cite{zhang2024the} demonstrates that  this similarity is unique to diffusion and flow matching and does not occur in other generative models such as GANs or VAEs.

Thus, there is a unique and notable similarity between the two mappings, which aligns with our analysis and motivates our diffusion-guided coupling approach to enhance flow matching.
Thanks to abundant existing diffusion models, flow matching with a diffusion-based coupling prior will be a convenient choice, simplifying coupling discovery and generating samples with fewer steps and straighter trajectories.

\subsection{Diffusion Model As a Coupling Prior}\label{subsec:straightfm}
Identifying effective couplings between image and noise is key to the success of previous methods.  
We aim to build straight trajectories for flow matching with coupling guidance across two distributions to facilitate training and shorten sampling steps.
 The underlying connection between flow matching and the PF-ODE of diffusion models allows one to use couplings provided by diffusion models to improve image quality and sampling process.

 Specifically, utilizing a well-trained diffusion model, we can generate one coupling
 $(\mathbf{x}_0,\tilde{\mathbf{x}}_1)$, by starting with an initial noise sample $\mathbf{x}_0\sim p_0$.
  Subsequently, $\tilde{\mathbf{x}}_1$ is sampled with one numerical ODE solver  along the PF-ODE trajectories of a diffusion model according to \cref{eq-pfode}.
After that, the coupling of noise and image samples $(\mathbf{x}_0,\tilde{\mathbf{x}}_1)$ is provided to optimize the subsequent objective:
\begin{equation}
        \mathcal{L}_{\text{revs}}(u_{\theta},\mathbf{x}_0,\tilde{\mathbf{x}}_1) \label{eq5}
=\mathbb{E}_{t,\mathbf{x}_0,\tilde{\mathbf{x}}_1}\|u_{\theta}(\mathbf x_t;t)- (\tilde{\mathbf{x}}_1-\mathbf{x}_0)\|^2,
\end{equation}
where  noise sample $ \mathbf{x}_{0}$ serves as the initial point for sampling a pseudo-image $\tilde{\mathbf{x}}_1$ from the PF-ODE  of diffusion model  and intermediate state $\mathbf x_{t}=t \tilde{\mathbf x}_{1}+(1-t) \mathbf x_{0}, \text{ for all } t \in[0,1]$.
As the coupling process is constructed from the reverse diffusion process, i.e., from noise to image, we use the shorthand notion  $\mathcal{L}_{\text{revs}}(u_{\theta},\mathbf{x}_0,\tilde{\mathbf{x}}_1)$ for ~\cref{eq5}.

Intuitively, the objective sets $(\tilde{\mathbf{x}}_1-\mathbf{x}_0)$  as the target velocity field $v$ in \cref{bcg-eq1}, representing the shortest straight path from noise to synthesized samples.
And the linear interpolation $\mathbf{x}_t$ can be formulated as the intermediate state at $t$ along this shortest path.
Following this,~\cref{eq5} is  derived by  plugging target velocity field $(\tilde{\mathbf{x}}_1-\mathbf{x}_0)$ into \cref{bcg-eq3cfm}.
After training, one can discretize ODE in \cref{bcg-eq1} just using N-step Euler method to generate samples as follows:
\begin{equation}\label{eq-sampling}
   \hat{\mathbf{x}}_{t_{i+1}} = \hat{\mathbf{x}}_{t_{i}}+(t_{i+1}-t_{i}) u_{\theta}\left(\hat{\mathbf{x}}_{t_{i}},t_i\right),
\end{equation}
where $\{t_i\}_{i=0}^{N-1}$ is an increasing sequence from 0 to 1.

Using samples generated by diffusion is appropriate due to the remarkable image quality achieved by contemporary diffusion models~\cite{Karras2022edm,rombach2022high},  which closely resembles real images.
Moreover, the coupling formed by diffusion model is also the approximation of optimal transport~\cite{khrulkov2023understanding}, which coincides with flow matching~\cite{liu2022flow,lipman2022flow}.
While Rectified Flow (RF) requires extensive multi-round training to generate successive couplings, such as using $k$-RF to synthesize images for $k+1$-RF (where $k \geq 2$), StraightFM achieves the same goal in a single round of synthetic data generation without requiring iterative multi-round training. This approach skips the potential degradation in image quality or diversity.

\subsection{Expanding Version: {Combined} with Real Samples}\label{subsec:expandingv}
 On a parallel track, existing works explore effective coupling direction from real data $\mathbf x_1\sim p_1$ to noise $\mathbf x_0\sim p_0$, e.g., Fast-ode~\cite{lee2023minimizing}, which utilizes an encoder to find couplings.
 However, there is still the opportunity to improve its empirical performance in few-step generation.
 Fast-ode can integrate with StraightFM, and then the coupling process is derived from two directions: noise to image and image to noise.
Concretely, we utilize the neural network $q_\phi(\tilde{\mathbf{x}}_0|\mathbf{x}_1)$ with parameters $\phi$ 	to represent the coupling process from real samples to noise samples, where we define $q_\phi(\tilde{\mathbf{x}}_0|\mathbf{x}_1)$ as a Gaussian distribution like $p_0$. 
Then, we optimize the following objective to ensure the validity of the coupling  ($\tilde{\mathbf{x}}_0,\mathbf{x}_1$): $  \mathcal{D}_{KL}(q_{\phi}(\tilde{\mathbf{x}}_0|\mathbf{x}_1)\|p_0)= \frac{1}{2}\sum_{i=1}^d ( \mu_{(i)}^2 +\sigma^2_{(i)} - \log \sigma^2_{(i)}-1 )$.
Couplings from image to noise are provided to optimize the following objective:
\begin{equation}
        \label{eq-forward}
 \mathcal{L}_{\text{forw}}(u_{\theta},\tilde{\mathbf{x}}_0,\mathbf{x}_1) 
=\mathbb{E}_{t,\tilde{\mathbf{x}}_0,\mathbf{x}_1}\|u_{\theta}(\mathbf x_t;t)- (\mathbf{x}_1-\tilde{\mathbf{x}}_0)\|^2.
\end{equation}
Here, $\mathbf x_t$ is also denoted as the linear interpolation between the coupling $(\tilde{\mathbf{x}}_0,\mathbf{x}_1)$.
Corresponding to \cref{eq5}, we use the notation $\mathcal{L}_{\text{forw}}$ for this objective.
The overall objective is formulated as:
\begin{equation}
         \label{eq-straightFM}
\min_{\theta,\phi } \mathcal{L}_{\text{revs}}(u_{\theta},\mathbf{x}_0,\tilde{\mathbf{x}}_1)+\lambda \mathcal{D}_{KL}(q_\phi (\tilde{\mathbf{x}}_0|\mathbf x_{1})\| p_0)+
\mathcal{L}_{\text{forw}}(u_{\theta},\tilde{\mathbf{x}}_0,\mathbf{x}_1),  
\end{equation}
where $\lambda$ is a hyperparameter.
We can jointly minimize \cref{eq-straightFM}  to update the parameters of $u_{\theta}$ and $q_{\phi}$ with a fast convergence speed.
{Attributed to the coupling from two complementary directions, this expanding version can obtain the capabilities of few-step generation through  \cref{eq-sampling}.}

\subsection{Compatible with Guided Conditional Generation}\label{subsec:mmgen}
As a fast-sampling unconditional generative model deriving from flow matching, StraightFM can cooperate with multimodal guided sampling to achieve conditional generation without any retraining-specific components.
 Inspired by a training-free multimodal guided sampling method designed for diffusion models~\cite{yu2023freedom}, we get a general conditional sampling process for StraightFM:
\begin{equation}\label{eq:condsampling}
    \hat{\mathbf{x}}_{t_{i+1}}=\hat{\mathbf{x}}_{t_{i}}+(t_{i+1}-t_{i}) u_{\theta}\left(\hat{\mathbf{x}}_{t_{i}},t_i\right)-\rho_t \nabla_{\hat{\mathbf{x}}_{t_{i}}} {E}\left(\mathbf{c},\hat{\mathbf{x}}_{t_{i}}\right).
\end{equation}
\cref{eq:condsampling} is different from the unconditional sampling process in \cref{eq-sampling}. 
Here, $\rho_t$ is a scale factor, and
$E$ is an energy function that measures the compatibility between the multimodal condition \(\mathbf c\) and the generated image, usually approximated as: 
% 
% $E$ is approximated as:
\begin{equation}\label{eq-energy}
  {E}\left(\mathbf{c},\hat{\mathbf{x}}_{t_{i}}\right)\approx \mathcal{D}\left(\mathcal{P}\left(\mathbf{c}\right),\mathcal{P}\left(\hat{\mathbf{x}}_{1|t_i}\left(\hat{\mathbf{x}}_{t_{i}}\right)\right)\right),
\end{equation}
where $\hat{\mathbf{x}}_{t_{i}}$ is employed to estimate the clean image  $\hat{\mathbf{x}}_{1|t_i}$,  formulated as: $ \hat{\mathbf{x}}_{1|t_i}= \hat{\mathbf{x}}_{t_{i}}+(1-t_i)u_{\theta}(\hat{\mathbf{x}}_{t_{i}},t_i)$, and the measure $ \mathcal{D}$  quantifies compatibility using 
  an off-the-shelf, pre-trained model   $\mathcal{P}$, which projects condition $\mathbf{c}$ and $\hat{\mathbf{x}}_{1|t_i}$ into a task-specific representation space.

% \vspace{-2mm}
\section{Experiments}\label{sec:exp}

In this section, we justify the advantages of  StraightFM  and the expanding version combined with real data~\cite{lee2023minimizing} (abbreviated as \textit{Fast-ode+Ours} in the remainder of this work) for image generation task. 
We further apply our unconditional generative model to various multimodal conditional generation tasks.
Experiments show that our proposed method yields high-quality samples within fewer sampling steps.
The ablation study, additional results, and implementation details are provided in the Supplementary Material.

\subsection{Image Generation on Pixel Space}\label{subsec:cifar10} 
\subsubsection{{Settings}}
In our experiments on CIFAR-10, we employ diffusion model EDM~\cite{Karras2022edm} with 17 Heun's 2nd-order steps as a coupling prior for StraightFM.
 We follow the plain parameters as RF~\cite{liu2022flow}, with images of 32 $\times$32 resolution.
 For the ODE solver, we test StraightFM on Euler method with uniform fixed steps and Runge-Kutta method of order 5(4) with adaptive steps.
 Performance is assessed according to IS (Inception Score) and FID (Fr\'echet Inception Distance).
The percentage of couplings from two directions of Fast-ode+Ours is 50\%, and we leave exploring different data percentages as future work.
 Details in  Supp.~1.1.

\subsubsection{{Baselines}} StraightFM is compared to several competitive baselines on CIFAR-10 dataset. 
These baselines are grouped into four categories. 
 (1) \textit{Flow-based models}: Rectified Flow~\cite{liu2022flow}, Fast-ode~\cite{lee2023minimizing}, FM~\cite{lipman2022flow}, I-CFM~\cite{tong2023improving}, and OT-CFM~\cite{tong2023improving}.
(2) \textit{Diffusion models}:  ScoreSDE~\cite{song2021score}, diffusion-GAN~\cite{xiao2022tackling},  EDM~\cite{Karras2022edm}, and Consistency Model~\cite{song2023consistency}. Here, diffusion-GAN and Consistency Model have the ability to one-step generation.
(3) \textit{Distillation-based models} Knowledge Distillation (KD)~\cite{luhman2021knowledge} and Progressive Distillation (PD)~\cite{salimans2022progressive}.
(4) \textit{Accelerated solver}: DPM-solver~\cite{lu2022dpm}, DEIS~\cite{zhang2022deis}, and UniPC~\cite{zhao2023unipc} reduce sampling steps to 5$\sim$20 without training.
 AMED-solver~\cite{zhou2023fastamed}, DM-NonUniform~\cite{xue2024accelerating}, and Bellman Steps~\cite{nguyen2024bellman} also  are accelerated methods focusing on optimizing time steps ($<$5 steps).
 \vspace{-0.15cm}

\begin{figure*}[htp]
\begin{minipage}{0.475\textwidth}
    \centering
    \captionof{table}{Sample quality assessment of different steps (N) on CIFAR-10, evaluated using IS and FID-50K. $\uparrow$ means higher is better, and $\downarrow$ means lower is better. \textbf{Bold} indicates the best result. {\ul Underline} indicates the second-best.
}
  \renewcommand{\arraystretch}{0.55}
   \resizebox{1\textwidth}{!}{
\begin{tabular}{@{}l|l|cr@{}}
\toprule
Sampling Steps & Method & IS$\uparrow$ & FID$\downarrow$ \\ \midrule
  N=5  & AMED-Solver~\cite{zhou2023fastamed} & {-} & 17.94  \\
   & EDM~\cite{Karras2022edm} & 7.81 & 35.51 \\
 % &DPM-solver ~\cite{lu2022dpm} & - & >50 \\
 % &3-DEIS~\cite{zhang2022deis} & - & 15.37 \\
  & UniPC+DM-NonUniform~\cite{xue2024accelerating} & {-} & 12.11 \\
    & RF~\cite{liu2022flow}+Bellman Steps ~\cite{nguyen2024bellman} & {7.22} & {32.70} \\
     & {Fast-ode~\cite{lee2023minimizing}} & 7.69 & 25.97 \\
 & \textbf{Fast-ode+Ours}  & {\ul 8.65} & {\ul 8.15} \\
 & \textbf{StraightFM (Ours)} & \textbf{9.28} & \textbf{3.25} \\
 \midrule
N=1 & 1-Rectified Flow~\cite{liu2022flow} & 1.13 & 378 \\
& 2-Rectified Flow~\cite{liu2022flow} & 8.08 & 12.21 \\
 & 3-Rectified Flow~\cite{liu2022flow} & 8.47 & \textbf{8.15} \\
 & FM~\cite{lipman2022flow} & - & 362.4 \\
  & I-CFM~\cite{tong2023improving} & - & 360.8 \\
 & OT-CFM~\cite{tong2023improving} & - & {239.0} \\
 & Consistency Model (CT)~\cite{song2023consistency} & {\ul 8.49} & 8.70 \\
  &  Progressive Distillation (PD)~\cite{salimans2022progressive} & - & 9.12  \\
 & Knowledge Distillation (KD)~\cite{luhman2021knowledge} & - & 9.36 \\
 & diffusion-GAN~\cite{xiao2022tackling} & 5.27  & 41.44 \\
 & Fast-ode~\cite{lee2023minimizing} & - & 289.0 \\
 & \textbf{Fast-ode+Ours} & 7.41 & 29.40 \\ 
 & \textbf{StraightFM (Ours)} & {\textbf{8.56}} & {\ul 8.65} \\
 \bottomrule
\end{tabular}%
}
\label{tab:cifar10}
\end{minipage}
\begin{minipage}{0.49\textwidth}
	\begin{subfloat}
			\centering     \includegraphics[scale=0.675]{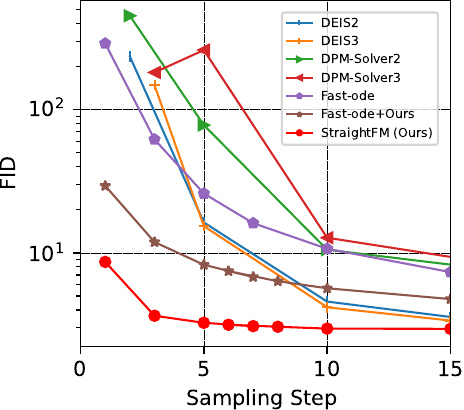}
		\end{subfloat}
    \caption{FID performance across different samplers and sampling  steps on CIFAR-10.  Fast-ode, Fast-ode+Ours and StraightFM still employ Euler method for sampling.}
    \label{fig:fid-nfe}\end{minipage}    \hfill
\end{figure*}
\vspace{-0.5cm}

\subsubsection{{Quantitative Comparison}}
 As shown in \cref{tab:cifar10}, the quality and diversity of images are evaluated using IS and FID.
Besides, \cref{fig:fid-nfe} displays FID performance across various sampling steps compared with existing popular diffusion samplers~\cite{zhang2022deis,lu2022dpm}.
Specifically, StraightFM is on par with, or better than, most baselines in one-step generation and outperforms most competitive baselines in  few-step generation. 
 At small steps (N $<$3), FID scores from StraightFM are nearly an order of magnitude lower than other methods as shown in~\cref{fig:fid-nfe}, indicating superior image quality.
As the sampling step (N) increases, StraightFM maintains image quality comparable to popular diffusion models with accelerated solvers. 
Fast-ode~\cite{lee2023minimizing}, which relies solely on the image-to-noise coupling process, fails in  few-step generation. 
By introducing a diffusion model as a coupling prior, it achieves significant improvements in few-step generation.

\newcommand{\scalefactorfig}{0.8} 
\begin{figure}[!thbp]
% \hspace{3mm}
\centering

	\hspace{3mm}
\raisebox{2.5mm}[0pt][0pt]{\begin{subfloat}
{\rotatebox{90}{\tiny EDM}}
\end{subfloat}}\hspace{-0.5mm}
	\begin{subfloat}
			\centering	\includegraphics[scale=\scalefactorfig]{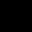}
	\end{subfloat}\hspace{-1.5mm}
		\begin{subfloat}
			\centering	\includegraphics[scale=\scalefactorfig]{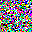}
	\end{subfloat}\hspace{-1.5mm}
		\begin{subfloat}
			\centering	\includegraphics[scale=\scalefactorfig]{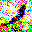}
	\end{subfloat}\hspace{-1.5mm}
		\begin{subfloat}
			\centering	\includegraphics[scale=\scalefactorfig]{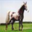}
	\end{subfloat}	\hspace{0.5mm}	
	\begin{subfloat}
			\centering	\includegraphics[scale=\scalefactorfig]{fig/2/edm/nfe1/0.png}
	\end{subfloat}\hspace{-1.5mm}
		\begin{subfloat}
			\centering	\includegraphics[scale=\scalefactorfig]{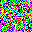}
	\end{subfloat}\hspace{-1.5mm}
		\begin{subfloat}
			\centering	\includegraphics[scale=\scalefactorfig]{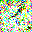}
	\end{subfloat}\hspace{-1.5mm}
		\begin{subfloat}
			\centering	\includegraphics[scale=\scalefactorfig]{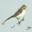}
	\end{subfloat}

	\hspace{3mm}
\raisebox{2.5mm}[0pt][0pt]{\begin{subfloat}
{\rotatebox{90}{\tiny 1-RF}}
\end{subfloat}}\hspace{-0.5mm}
	\begin{subfloat}
			\centering \includegraphics[scale=\scalefactorfig]{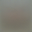}
	\end{subfloat}\hspace{-1.5mm}
		\begin{subfloat}
			\centering \includegraphics[scale=\scalefactorfig]{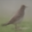}
	\end{subfloat}\hspace{-1.5mm}
		\begin{subfloat}
			\centering \includegraphics[scale=\scalefactorfig]{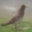}
	\end{subfloat}\hspace{-1.5mm}
	\begin{subfloat}
			\centering \includegraphics[scale=\scalefactorfig]{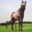}
	\end{subfloat}	\hspace{0.5mm}
		\begin{subfloat}
			\centering \includegraphics[scale=\scalefactorfig]{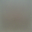}
	\end{subfloat}\hspace{-1.5mm}
		\begin{subfloat}
			\centering \includegraphics[scale=\scalefactorfig]{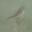}
	\end{subfloat}\hspace{-1.5mm}
		\begin{subfloat}
			\centering \includegraphics[scale=\scalefactorfig]{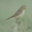}
	\end{subfloat}\hspace{-1.5mm}
		\begin{subfloat}
			\centering \includegraphics[scale=\scalefactorfig]{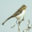}
	\end{subfloat}\hspace{0mm}

	\hspace{3mm}
\raisebox{2.5mm}[0pt][0pt]{\begin{subfloat}
{\rotatebox{90}{\tiny 2-RF}}
\end{subfloat}}\hspace{-0.5mm}
	\begin{subfloat}
			\centering \includegraphics[scale=\scalefactorfig]{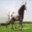}
	\end{subfloat}\hspace{-1.5mm}
		\begin{subfloat}
			\centering \includegraphics[scale=\scalefactorfig]{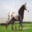}
	\end{subfloat}\hspace{-1.5mm}
		\begin{subfloat}
			\centering \includegraphics[scale=\scalefactorfig]{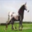}
	\end{subfloat}\hspace{-1.5mm}
		\begin{subfloat}
			\centering \includegraphics[scale=\scalefactorfig]{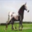}
	\end{subfloat}	\hspace{0.5mm}
		\begin{subfloat}
			\centering \includegraphics[scale=\scalefactorfig]{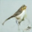}
	\end{subfloat}\hspace{-1.5mm}
		\begin{subfloat}
			\centering \includegraphics[scale=\scalefactorfig]{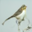}
	\end{subfloat}\hspace{-1.5mm}
	\begin{subfloat}
			\centering \includegraphics[scale=\scalefactorfig]{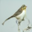}
	\end{subfloat}\hspace{-1.5mm}
		\begin{subfloat}
			\centering \includegraphics[scale=\scalefactorfig]{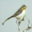}
	\end{subfloat}

	\hspace{3mm}
\raisebox{2.25mm}[0pt][0pt]{\begin{subfloat}
{\rotatebox{90}{\hspace{2.5mm}\tiny 3-RF}}
\end{subfloat}}\hspace{-0.5mm}
	\begin{subfloat}
			\centering \includegraphics[scale=\scalefactorfig]{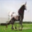}
	\end{subfloat}\hspace{-1.5mm}
		\begin{subfloat}
			\centering \includegraphics[scale=\scalefactorfig]{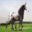}
	\end{subfloat}\hspace{-1.5mm}
		\begin{subfloat}
			\centering \includegraphics[scale=\scalefactorfig]{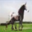}
	\end{subfloat}\hspace{-1.5mm}
		\begin{subfloat}
			\centering \includegraphics[scale=\scalefactorfig]{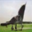}
	\end{subfloat}	\hspace{0.5mm}
		\begin{subfloat}
			\centering \includegraphics[scale=\scalefactorfig]{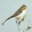}
	\end{subfloat}\hspace{-1.5mm}
		\begin{subfloat}
			\centering \includegraphics[scale=\scalefactorfig]{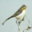}
	\end{subfloat}\hspace{-1.5mm}
		\begin{subfloat}
			\centering \includegraphics[scale=\scalefactorfig]{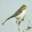}
	\end{subfloat}\hspace{-1.5mm}
		\begin{subfloat}
			\centering \includegraphics[scale=\scalefactorfig]{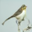}
	\end{subfloat}

	\hspace{2.75mm}
\raisebox{1.1mm}[0pt][0pt]{\begin{subfloat}
{\rotatebox{90}{\hspace{-3.5mm}\tiny Fast-ode+Ours}}\end{subfloat}}\hspace{-0.5mm}
	\begin{subfloat}
			\centering \includegraphics[scale=\scalefactorfig]{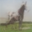}
	\end{subfloat}\hspace{-1.5mm}
		\begin{subfloat}
			\centering \includegraphics[scale=\scalefactorfig]{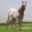}
	\end{subfloat}\hspace{-1.5mm}
		\begin{subfloat}
			\centering \includegraphics[scale=\scalefactorfig]{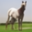}
	\end{subfloat}\hspace{-1.5mm}
		\begin{subfloat}
			\centering \includegraphics[scale=\scalefactorfig]{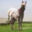}
	\end{subfloat}	\hspace{0.5mm}
		\begin{subfloat}
			\centering \includegraphics[scale=\scalefactorfig]{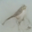}
	\end{subfloat}\hspace{-1.5mm}
			\begin{subfloat}
			\centering \includegraphics[scale=\scalefactorfig]{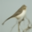}
	\end{subfloat}\hspace{-1.5mm}
			\begin{subfloat}
			\centering \includegraphics[scale=\scalefactorfig]{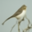}
	\end{subfloat}\hspace{-1.5mm}
			\begin{subfloat}
			\centering \includegraphics[scale=\scalefactorfig]{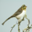}
	\end{subfloat}\hspace{0mm}

	\hspace{4.125mm}
\raisebox{1.0mm}[0pt][0pt]{\begin{subfloat}
{\rotatebox{90}{\tiny Ours \hspace{-7.5mm}}}
\end{subfloat}}\hspace{-0.5mm}
	\begin{subfloat}
			\centering \includegraphics[scale=\scalefactorfig]{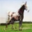}
	\end{subfloat}\hspace{-1.5mm}
		\begin{subfloat}
			\centering \includegraphics[scale=\scalefactorfig]{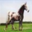}
	\end{subfloat}\hspace{-1.5mm}
			\begin{subfloat}
			\centering \includegraphics[scale=\scalefactorfig]{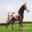}
	\end{subfloat}\hspace{-1.5mm}
			\begin{subfloat}
			\centering \includegraphics[scale=\scalefactorfig]{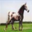}
	\end{subfloat}	\hspace{0.5mm}
			\begin{subfloat}
			\centering \includegraphics[scale=\scalefactorfig]{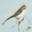}
	\end{subfloat}\hspace{-1.5mm}
					\begin{subfloat}
			\centering \includegraphics[scale=\scalefactorfig]{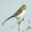}
	\end{subfloat}\hspace{-1.5mm}
					\begin{subfloat}
			\centering \includegraphics[scale=\scalefactorfig]{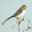}
	\end{subfloat}\hspace{-1.5mm}
					\begin{subfloat}
			\centering \includegraphics[scale=\scalefactorfig]{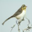}
		\end{subfloat}
 \vspace{-2mm}

  \hspace{3mm}
		\begin{minipage}[h]{0.8cm}
			\centering {\tiny N=1}
		\end{minipage}\hspace{0.7mm}
		\begin{minipage}[h]{0.8cm}
			\centering  {\tiny N=3}
		\end{minipage}\hspace{0.75mm}
		\begin{minipage}[h]{0.8cm}
			\centering  {\tiny N=5}
		\end{minipage}\hspace{0.2mm}
		\begin{minipage}[h]{0.8cm}
			\centering  {\tiny Ada.N}
		\end{minipage}\hspace{4mm}
  		\begin{minipage}[h]{0.8cm}
			\centering {\tiny N=1}
		\end{minipage}\hspace{0.5mm}
		\begin{minipage}[h]{0.8cm}
			\centering  {\tiny N=3}
		\end{minipage}\hspace{0.5mm}
		\begin{minipage}[h]{0.8cm}
			\centering  {\tiny N=5}
		\end{minipage}\hspace{0.5mm}
		\begin{minipage}[h]{0.8cm}
			\centering  {\tiny Ada.N}
		\end{minipage}
    \caption{Comparison includes different models trained on CIFAR-10 when simulated with fixed and adaptive steps (abbreviated as N). All corresponding samples are generated from the same initial noise. EDM uses 2nd order Heun method with 17 steps  in adaptive sampling.}
    \label{fig:cifar10}
\end{figure}

\subsubsection{{Qualitative Comparison}}
Figure \ref{fig:cifar10} illustrates generated images  with varying sampling steps. 
 Leveraging diffusion models as the coupling prior, StraightFM and Fast-ode+Ours consistently produce high-quality images, largely independent of the sampling step.
However, the generated results of EDM and 1-RF are highly sensitive to the sampling step $N$ (as seen in the 3rd and 4th columns of 1-RF and EDM in \cref{fig:cifar10}).
This suggests that our trajectories are straighter than the other models.
Moreover, StraightFM and 1-RF undertake single-round training, but only StraightFM can generate images in a single step. 
Despite multi-round training, 3-RF still shows  obvious distinction when changing the sampling step.
 Importantly, StraightFM's outputs differ from EDM's, even with shared couplings, indicating it captures the entire image distribution rather than simply memorizing training data.

\subsection{Image Generation on Latent Space}\label{subsec:celebahq}
\subsubsection{{Settings} }
Conducting the generative process in the latent space~\cite{rombach2022high} provides a different approach to high-resolution image generation with lower computational costs.
Thus, we apply StraightFM in the latent space of the pre-trained VAE from Latent Diffusion Model (LDM)~\cite{rombach2022high} on CelebA-HQ 256 $\times$ 256, improving computational efficiency in high-resolution image synthesis.
LDM serves as a coupling prior for StraightFM training, utilizing same hyper-parameters from Latent Flow Matching (LFM)~\cite{dao2023flow} with DiT~\cite{Peebles2022DiT}. 
For coupling sampling, we adopt DDIM sampler~\cite{song2020ddim},  a specific instance of PF-ODE~\cite{song2020ddim}.
 Results are evaluated using FID (Fr\'echet Inception Distance) and KID (Kernel Inception Distance). Details in Supp.~1.2.

\vspace{-0.25cm}
\subsubsection{{Baselines} }
\begin{figure*}[!thbp]
\hspace{6mm}\raisebox{0.55cm}[0pt][0pt]{\begin{subfloat}
{\rotatebox{90}{N=5}}
\end{subfloat}}\hspace{-0.5mm}
	\begin{subfloat}
			\centering \includegraphics[scale=0.22]{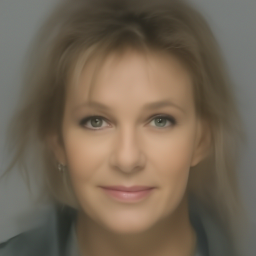}
		\end{subfloat}\hspace{-1.5mm}
	\begin{subfloat}
			\centering \includegraphics[scale=0.22]{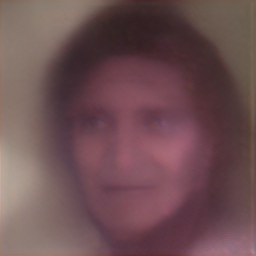}
		\end{subfloat}\hspace{-1.5mm}
	\begin{subfloat}
			\centering \includegraphics[scale=0.22]{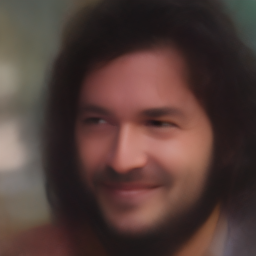}
		\end{subfloat}\hspace{-1.5mm}
			\begin{subfloat}
			\centering \includegraphics[scale=0.22]{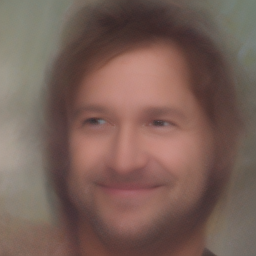}
		\end{subfloat}\hspace{-1.5mm}
			\begin{subfloat}
			\centering \includegraphics[scale=0.22]{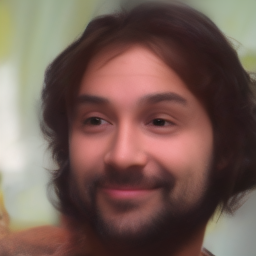}
		\end{subfloat}

\hspace{6mm}\raisebox{0.55cm}[0pt][0pt]{\begin{subfloat}
{\rotatebox{90}{N=10}}
\end{subfloat}}\hspace{-0.5mm}	\begin{subfloat}
			\centering \includegraphics[scale=0.22]{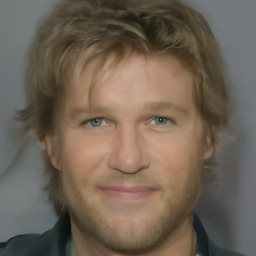}
		\end{subfloat}\hspace{-1.5mm}
   	\begin{subfloat}
			\centering \includegraphics[scale=0.22]{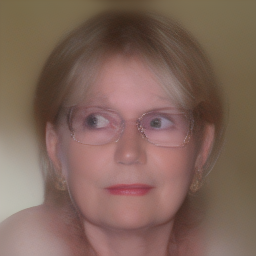}
		\end{subfloat}\hspace{-1.5mm}
   	\begin{subfloat}
			\centering \includegraphics[scale=0.22]{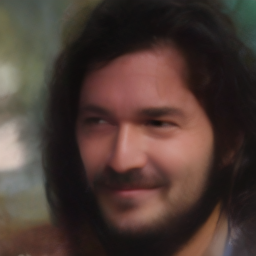}
		\end{subfloat}\hspace{-1.5mm}
   	\begin{subfloat}
			\centering \includegraphics[scale=0.22]{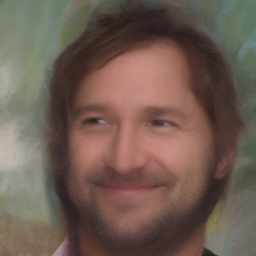}
		\end{subfloat}\hspace{-1.5mm}
   	\begin{subfloat}
			\centering \includegraphics[scale=0.22]{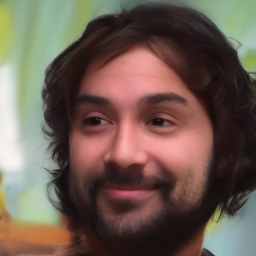}
   \end{subfloat}

   	\hspace{6mm}\raisebox{0.55cm}[0pt][0pt]{\begin{subfloat}
{\rotatebox{90}{N=15}}
\end{subfloat}}\hspace{-0.5mm}
	\begin{subfloat}
			\centering \includegraphics[scale=0.22]{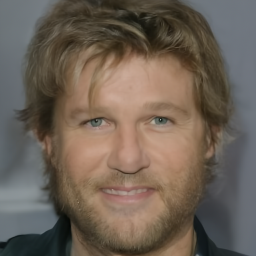}
		\end{subfloat}\hspace{-1.5mm}
		\begin{subfloat}
			\centering \includegraphics[scale=0.22]{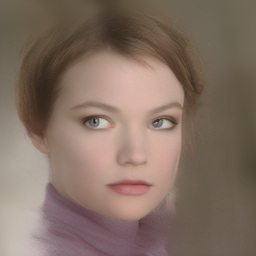}
		\end{subfloat}\hspace{-1.5mm}
			\begin{subfloat}
			\centering \includegraphics[scale=0.22]{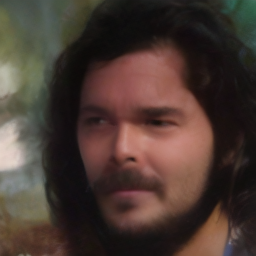}
		\end{subfloat}\hspace{-1.5mm}
			\begin{subfloat}
			\centering \includegraphics[scale=0.22]{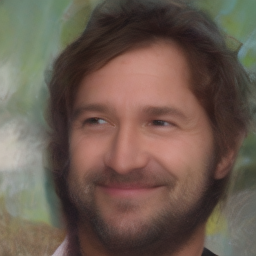}
		\end{subfloat}\hspace{-1.5mm}
			\begin{subfloat}
			\centering \includegraphics[scale=0.22]{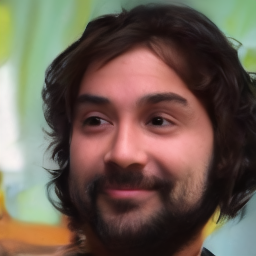}
		\end{subfloat} 
		
		\hspace{6mm}\raisebox{0.55cm}[0pt][0pt]{\begin{subfloat}
{\rotatebox{90}{N=20}}
\end{subfloat}}\hspace{-0.5mm}
	\begin{subfloat}
			\centering \includegraphics[scale=0.22]{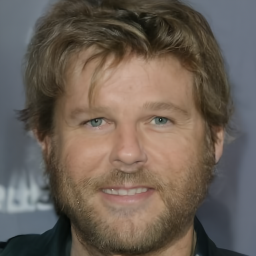}
		\end{subfloat}\hspace{-1.5mm}
			\begin{subfloat}
			\centering \includegraphics[scale=0.22]{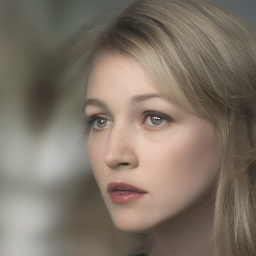}
		\end{subfloat}\hspace{-1.5mm}
		\begin{subfloat}
			\centering \includegraphics[scale=0.22]{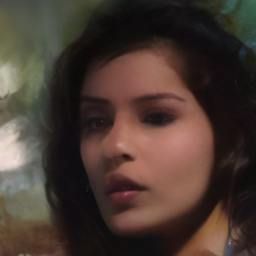}
		\end{subfloat}\hspace{-1.5mm}
			\begin{subfloat}
			\centering \includegraphics[scale=0.22]{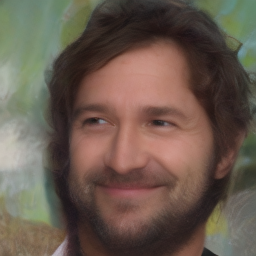}
		\end{subfloat}\hspace{-1.5mm}
	\begin{subfloat}
			\centering \includegraphics[scale=0.22]{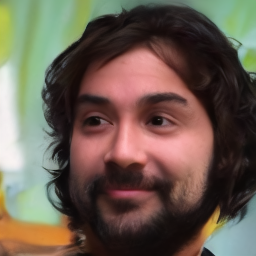}
		\end{subfloat} \vspace{0.5mm}		
		
\hspace{6mm}\raisebox{0.55cm}[0pt][0pt]{\begin{subfloat}
{\rotatebox{90}{N=5}}
\end{subfloat}}\hspace{-0.5mm}	\begin{subfloat}
			\centering \includegraphics[scale=0.22]{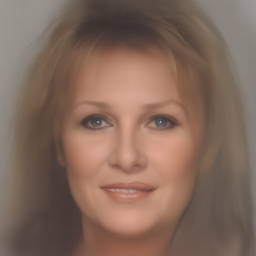}
		\end{subfloat}\hspace{-1.5mm}
	\begin{subfloat}
			\centering \includegraphics[scale=0.22]{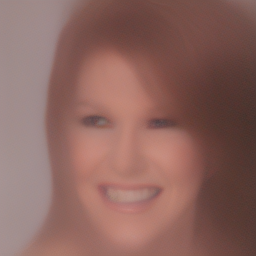}
		\end{subfloat}\hspace{-1.5mm}
	\begin{subfloat}
			\centering \includegraphics[scale=0.22]{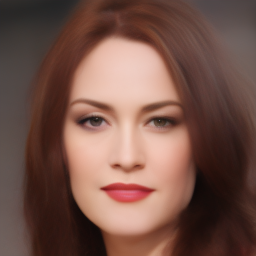}
		\end{subfloat}\hspace{-1.5mm}
	\begin{subfloat}
			\centering \includegraphics[scale=0.22]{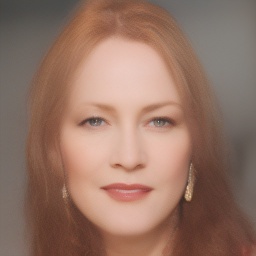}
		\end{subfloat}\hspace{-1.5mm}
	\begin{subfloat}
			\centering \includegraphics[scale=0.22]{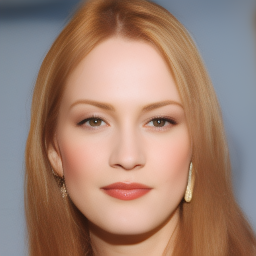}
		\end{subfloat} 		
		
		\hspace{6mm}\raisebox{0.55cm}[0pt][0pt]{\begin{subfloat}
{\rotatebox{90}{N=10}}
\end{subfloat}}\hspace{-0.5mm}
	\begin{subfloat}
			\centering \includegraphics[scale=0.22]{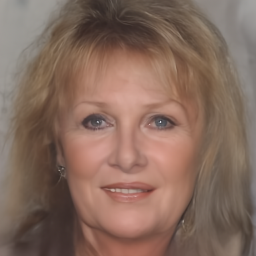}
		\end{subfloat}\hspace{-1.5mm}
	\begin{subfloat}
			\centering \includegraphics[scale=0.22]{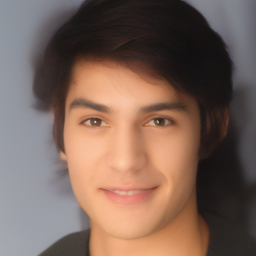}
		\end{subfloat}\hspace{-1.5mm}
	\begin{subfloat}
			\centering \includegraphics[scale=0.22]{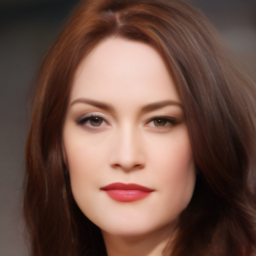}
		\end{subfloat}\hspace{-1.5mm}
		\begin{subfloat}
			\centering \includegraphics[scale=0.22]{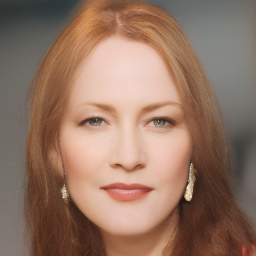}
		\end{subfloat}\hspace{-1.5mm}
	\begin{subfloat}
			\centering \includegraphics[scale=0.22]{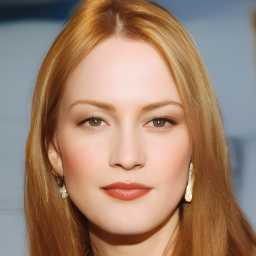}
		\end{subfloat}

		\hspace{6mm}\raisebox{0.55cm}[0pt][0pt]{\begin{subfloat}
{\rotatebox{90}{N=15}}
\end{subfloat}}\hspace{-0.5mm}	
	\begin{subfloat}
			\centering \includegraphics[scale=0.22]{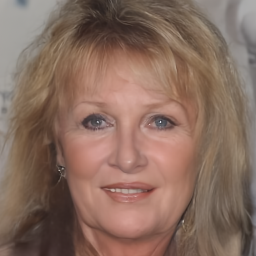}
		\end{subfloat}\hspace{-1.5mm}
	\begin{subfloat}
			\centering \includegraphics[scale=0.22]{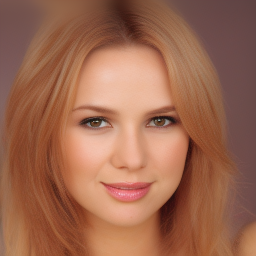}
		\end{subfloat}\hspace{-1.5mm}
	\begin{subfloat}
			\centering \includegraphics[scale=0.22]{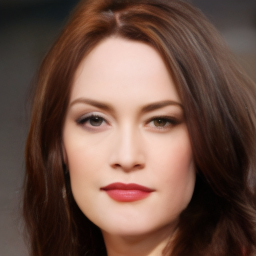}
		\end{subfloat}\hspace{-1.5mm}
	\begin{subfloat}
			\centering \includegraphics[scale=0.22]{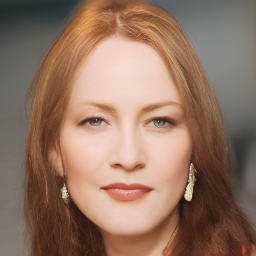}
		\end{subfloat}\hspace{-1.5mm}
	\begin{subfloat}
			\centering \includegraphics[scale=0.22]{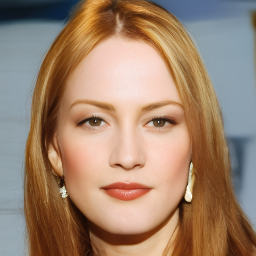}
		\end{subfloat} 		
		
	\hspace{6mm}\raisebox{0.55cm}[0pt][0pt]{\begin{subfloat}
{\rotatebox{90}{N=20}}
\end{subfloat}}\hspace{-0.5mm}	
  	\begin{subfloat}
			\centering \includegraphics[scale=0.22]{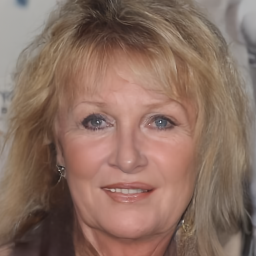}
		\end{subfloat}\hspace{-1.5mm}
	\begin{subfloat}
			\centering \includegraphics[scale=0.22]{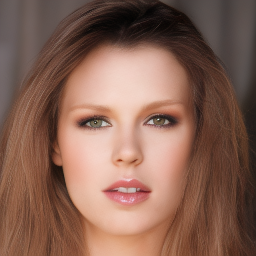}
		\end{subfloat}\hspace{-1.5mm}
  	\begin{subfloat}
			\centering \includegraphics[scale=0.22]{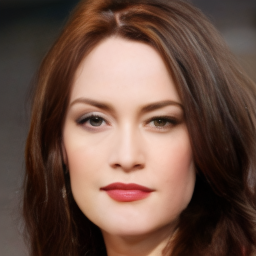}
		\end{subfloat}\hspace{-1.5mm}
  	\begin{subfloat}
			\centering \includegraphics[scale=0.22]{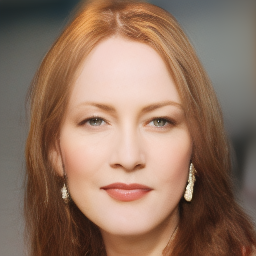}
		\end{subfloat}\hspace{-1.5mm}
 	\begin{subfloat}
			\centering \includegraphics[scale=0.22]{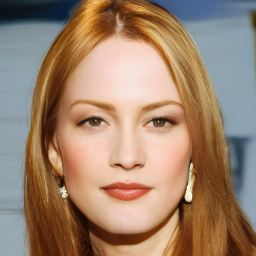}
		\end{subfloat} 	\vspace{-1mm}

	\hspace{2mm} \hspace{2mm}\begin{minipage}[t]{0.15\linewidth}
			\centering {\small RF}
		\end{minipage}%
  \hspace{2mm}
\begin{minipage}[t]{0.15\linewidth}
			\centering{\small LDM}
		\end{minipage}%
  \hspace{2mm}
    \begin{minipage}[t]{0.15\linewidth}
			\centering {\small LFM}
		\end{minipage}%
  \hspace{-1mm}
		\begin{minipage}[t]{0.2\linewidth}
			\centering {\small Fast-ode+Ours}
		\end{minipage}
  \hspace{1mm}
		\begin{minipage}[t]{0.15\linewidth}
			\centering {\small Ours }
		\end{minipage}    
    \caption{Samples generated on CelebA-HQ  dataset utilizing varying sampling steps (N). Here, corresponding images from LDM, LFM and StraightFM are initialized from the same random seed. Best viewed zoomed in.}
    \label{fig:celebahq}
\end{figure*}

 StraightFM is compared to Rectified Flow (RF)~\cite{liu2022flow}, Latent Flow Matching (LFM)~\cite{dao2023flow}, and Latent Diffusion Model (LDM)~\cite{rombach2022high} on CelebA-HQ dataset.
Results are presented in \cref{tab:celebahq} and \cref{fig:celebahq}.

\subsubsection{{Quantitative Comparison} }
The evaluation of FID and KID  for 10K generated images is reported in \cref{tab:celebahq}.
StraightFM consistently outperforms other competitive baselines on FID and KID metrics, achieving superior results with fewer steps.
Setting the sampling step to 3, FID and KID achieve improvements of approximately 19.8\%-37.1\% and 27.5\%-40.6\%, respectively, enhancing image quality as the sampling steps increase.
 When the sampling step is set to 20, FID and KID improve by 15.0\%-39.0\% and 22.6\%-44.9\%, respectively.
Moreover, StraightFM maintains high performance even with larger sampling steps.
Overall, StraightFM enables fast generation in the latent space.\vspace{-0.2cm}

% \vspace{-0.1cm}
\begin{table*}[!th]
\centering
\begin{minipage}{1\textwidth}
    \centering
\caption{Comparison of FID-10K (top) and KID-10K (bottom)  metrics on  CelebA-HQ 256$\times$ 256 dataset.   *denotes our own implementation. \textbf{Bold} indicates the best result. {\ul Underline} indicates the second-best.}\label{tab:celebahq}
\label{tab:celeba-fid} \renewcommand{\arraystretch}{0.9}
\resizebox{0.75\textwidth}{!}{
\begin{tabular}{@{}lccccccc@{}}
\toprule
Method$\backslash$  Step   & 3              & 5              & 10             & 15             & 20 &25     &30       \\ \midrule
Rectified Flow~\cite{liu2022flow} & 155.96         & 135.28         & 85.60          & 62.41          & 48.46 &    39.64 & 48.46     \\
Latent Diffusion Model~\cite{rombach2022high}       & 150.98         & 106.28         & 64.69          & 46.29          & 36.78     &  {\ul 31.50}   &{\ul 27.58}   \\
Latent Flow Matching*~\cite{dao2023flow}           & {\ul 122.11}         & {\ul 88.67}          & 55.64          & 42.52          & 36.25      &  32.90 &30.90   \\
Fast-ode~\cite{lee2023minimizing}+Ours  & 140.09   & 92.86    & {\ul 52.09}    & {\ul 40.10}    & {\ul 34.79}  & 31.62 & 29.75   \\
Ours           & \textbf{97.96} & \textbf{62.85} & \textbf{38.35} & \textbf{32.09} & \textbf{29.57} & \textbf{28.16} & \textbf{27.37} \\ \bottomrule
\end{tabular}%
}
\end{minipage}  

\begin{minipage}{0.75\textwidth}
\centering
\label{tab:celeba-kid}
\renewcommand{\arraystretch}{0.9}
\resizebox{\textwidth}{!}{
\begin{tabular}{@{}lccccccc@{}}
\toprule
Method$\backslash$ Step     & 3             & 5             & 10            & 15            & 20      & 25  &30    \\ \midrule
Rectified Flow~\cite{liu2022flow} & 0.156          & 0.147          & 0.090          & 0.064          & 0.049    &    0.039  &0.049  \\
Latent Diffusion Model~\cite{rombach2022high}           & 0.160          & 0.107          & 0.063          & 0.045          & 0.035     &  0.030 &{\ul 0.026}    \\
Latent Flow Matching*~\cite{dao2023flow}        & {\ul 0.131}    & 0.094          & 0.058          & 0.044          & 0.037      &  0.033 & 0.030   \\
Fast-ode\cite{lee2023minimizing}+Ours  & 0.139          & {\ul 0.089}    & {\ul 0.048}    & {\ul 0.036}    & {\ul 0.031} &    {\ul 0.028} & {\ul 0.026}  \\
Ours           & \textbf{0.095} & \textbf{0.059} & \textbf{0.036} & \textbf{0.030} & \textbf{0.027} & \textbf{0.026} & \textbf{0.025}  \\ \bottomrule
\end{tabular}%
}
\end{minipage}
\end{table*}
 \vspace{-0.25cm}

\vspace{-0.25cm}
\subsubsection{{Qualitative Comparison}}
The visualizations of sample trajectories can be found in \cref{fig:celebahq}.
LDM, LFM, and StraightFM use the same random seed on the latent space; RF is trained on pixel space. 
StraightFM generates more attractive images with straighter trajectories and is less constrained by the prior knowledge embedded in the diffusion model.
One clue is given by  \cref{fig:celebahq}, which shows that StraightFM is not constrained by the prior knowledge embedded in the diffusion model.
It is interesting that StraightFM, learning couplings from LDM, exhibits similar but more fine-grained generation compared to LFM learning in real data. 
 Moreover, StraightFM is highly independent of sampling steps, while generated images from LFM and LDM display unexpected attribute changes, e.g.,  the 2nd and 3rd rows in LFM.
This suggests that trajectories of StraightFM are straighter, enabling the generation of high-quality samples with fewer steps.

 \subsection{Multimodal Conditional Generation with Guided Sampling}\label{subsec:expmmgen}\vspace{-0.5mm}
Similar to a diffusion model based on training-free sampling to achieve conditional generation, unconditional StraightFM allows conditional generations with fewer steps.
% % 
We compare it to FreeDoM~\cite{yu2023freedom}, a general energy-guided conditional sampling method designed for diffusion models to demonstrate this capability.
  \begin{figure}[th]
			\centering \includegraphics[scale=0.375]{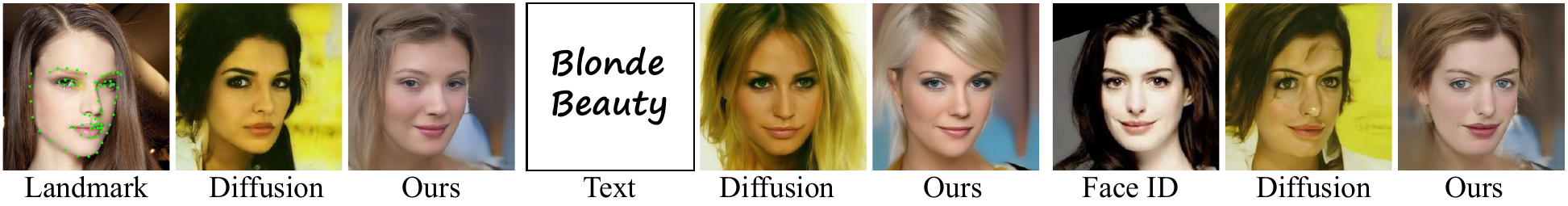}
     \caption{Training-free conditional image generation with various multimodal guidance, i.e., landmarks, text, and face ID, starting from a same random seed. Results from vanilla diffusion model~\cite{yu2023freedom} follows the default setting (N=50). The sampling step of  StraightFM  is set as 10.  }
     \label{fig:celebahq-freedom}
 \end{figure}

According to \cref{eq-energy}, the off-the-shelf pre-trained models for various condition are same as the default setting of FreeDoM.
 We set the scale factor  $\rho_t$ in \cref{eq:condsampling} as the signal-to-noise ratio of StraightFM, i.e., $\frac{1-t}{t}$, and initialization time is $t=0.2$.
Details in Supp.~1.3.
Results are displayed in \cref{fig:celebahq-freedom}. 
 StraightFM supports a wide range of conditional generation with $5\times$ speed up to achieve high-quality image generation, compared to the original diffusion model.

\section{Conclusion}\label{sec:conc}
{
Existing flow matching methods to reduce sampling step suffer from either multi-round training or  limited information within minibatches.
To address this issue, we propose StraightFM, which leverages diffusion model insights to enable efficient high-quality image generation and is compatible with training-free multimodal generation. 
{
Unlike traditional flow matching methods, StraightFM leverages information from an independent generative model, based on the unique similarity between flow matching and diffusion. 
This opens new possibilities for exploring connections among different ODE-based generative models. 
When dealing with more complex distributions, the absence of diffusion model guidance can pose challenges. 
Future studies focus on more tractable and effective coupling strategies in the generative community.
}}

\noindent
\subsubsection{\textbf{Acknowledgements.}}
This work was supported by the National Natural Science Foundation of China (NSFC) (Grant No. 62376265 \& 62576338), Young Scientists Fund of the State Key Laboratory of Multimodal Artificial Intelligence Systems (Grant No. ES2P100116), Beijing Natural Science Foundation (Grant No. L252145), and the Strategic Priority Research Program of Chinese Academy of Sciences (Grant No. Grant XDA0480302).

%

\begingroup
\sloppy
% \begin{raggedright}
\bibliographystyle{splncs04}
\bibliography{straightfm.bib}

\begin{thebibliography}{10}
\providecommand{\url}[1]{\texttt{#1}}
\providecommand{\urlprefix}{URL }
\providecommand{\doi}[1]{https://doi.org/#1}

\bibitem{albergo2023building}
Albergo, M.S., Vanden-Eijnden, E.: Building normalizing flows with stochastic interpolants. In: ICLR (2023)

\bibitem{alemohammad2023self}
Alemohammad, S., Casco-Rodriguez, J., Luzi, L., Humayun, A.I., Babaei, H., LeJeune, D., Siahkoohi, A., Baraniuk, R.G.: Self-consuming generative models go mad. In: ICLR (2024)

\bibitem{chen2018neural}
Chen, R.T., Rubanova, Y., Bettencourt, J., Duvenaud, D.K.: Neural ordinary differential equations. In: NeurIPS. vol.~31 (2018)

\bibitem{dao2023flow}
Dao, Q., Phung, H., Nguyen, B., Tran, A.: Flow matching in latent space. arXiv preprint arXiv:2307.08698  (2023)

\bibitem{de2021diffusion}
De~Bortoli, V., Thornton, J., Heng, J., Doucet, A.: Diffusion schr{\"o}dinger bridge with applications to score-based generative modeling. NeurIPS  \textbf{34},  17695--17709 (2021)

\bibitem{gao2025diffusionmeetsflow}
Gao, R., Hoogeboom, E., Heek, J., Bortoli, V.D., Murphy, K.P., Salimans, T.: Diffusion meets flow matching: Two sides of the same coin (2024), \url{https://diffusionflow.github.io/}

\bibitem{grathwohl2018ffjord}
Grathwohl, W., Chen, R.T., Bettencourt, J., Sutskever, I., Duvenaud, D.: Ffjord: Free-form continuous dynamics for scalable reversible generative models. In: ICLR (2018)

\bibitem{ho2020denoising}
Ho, J., Jain, A., Abbeel, P.: Denoising diffusion probabilistic models. In: NeurIPS. vol.~33, pp. 6840--6851 (2020)

\bibitem{Karras2022edm}
Karras, T., Aittala, M., Aila, T., Laine, S.: Elucidating the design space of diffusion-based generative models. NeurIPS  (2022)

\bibitem{khrulkov2023understanding}
Khrulkov, V., Ryzhakov, G., Chertkov, A., Oseledets, I.: Understanding ddpm latent codes through optimal transport. In: ICLR (2023)

\bibitem{kwon2022score}
Kwon, D., Fan, Y., Lee, K.: Score-based generative modeling secretly minimizes the wasserstein distance. In: NeurIPS. vol.~35, pp. 20205--20217 (2022)

\bibitem{lee2023minimizing}
Lee, S., Kim, B., Ye, J.C.: Minimizing trajectory curvature of ode-based generative models. In: ICML (2023)

\bibitem{lipman2022flow}
Lipman, Y., Chen, R.T., Ben-Hamu, H., Nickel, M., Le, M.: Flow matching for generative modeling. In: ICLR (2023)

\bibitem{liu2022flow}
Liu, X., Gong, C., et~al.: Flow straight and fast: Learning to generate and transfer data with rectified flow. In: ICLR (2023)

\bibitem{liu2023insta}
Liu, X., Zhang, X., Ma, J., Peng, J., Liu, Q.: Instaflow: One step is enough for high-quality diffusion-based text-to-image generation. In: ICLR (2024)

\bibitem{lu2022dpm}
Lu, C., Zhou, Y., Bao, F., Chen, J., Li, C., Zhu, J.: Dpm-solver: A fast ode solver for diffusion probabilistic model sampling in around 10 steps. In: NeurIPS. vol.~35, pp. 5775--5787 (2022)

\bibitem{luhman2021knowledge}
Luhman, E., Luhman, T.: Knowledge distillation in iterative generative models for improved sampling speed. arXiv preprint arXiv:2101.02388  (2021)

\bibitem{nguyen2024bellman}
Nguyen, B., Nguyen, B., Nguyen, V.A.: Bellman optimal step-size straightening of flow-matching models. In: ICLR (2024)

\bibitem{Peebles2022DiT}
Peebles, W., Xie, S.: Scalable diffusion models with transformers. In: ICCV (2023)

\bibitem{pooladian2023multisample}
Pooladian, A.A., Ben-Hamu, H., Domingo-Enrich, C., Amos, B., Lipman, Y., Chen, R.: Multisample flow matching: Straightening flows with minibatch couplings. In: ICML (2023)

\bibitem{rombach2022high}
Rombach, R., Blattmann, A., Lorenz, D., Esser, P., Ommer, B.: High-resolution image synthesis with latent diffusion models. In: CVPR. pp. 10684--10695 (2022)

\bibitem{salimans2022progressive}
Salimans, T., Ho, J.: Progressive distillation for fast sampling of diffusion models. In: ICLR (2022)

\bibitem{song2020ddim}
Song, J., Meng, C., Ermon, S.: Denoising diffusion implicit models. In: ICLR (2021)

\bibitem{song2023consistency}
Song, Y., Dhariwal, P., Chen, M., Sutskever, I.: Consistency models. In: ICML (2023)

\bibitem{song2021score}
Song, Y., Sohl-Dickstein, J., Kingma, D.P., Kumar, A., Ermon, S., Poole, B.: Score-based generative modeling through stochastic differential equations. In: ICLR (2021)

\bibitem{tong2023improving}
Tong, A., Malkin, N., Huguet, G., Zhang, Y., Rector-Brooks, J., Fatras, K., Wolf, G., Bengio, Y.: Improving and generalizing flow-based generative models with minibatch optimal transport. In: ICMLW (2023)

\bibitem{xiao2022tackling}
Xiao, Z., Kreis, K., Vahdat, A.: Tackling the generative learning trilemma with denoising diffusion gans. In: ICLR (2022)

\bibitem{xue2024accelerating}
Xue, S., Liu, Z., Chen, F., Zhang, S., Hu, T., Xie, E., Li, Z.: Accelerating diffusion sampling with optimized time steps. In: CVPR (2024)

\bibitem{yu2023freedom}
Yu, J., Wang, Y., Zhao, C., Ghanem, B., Zhang, J.: Freedom: Training-free energy-guided conditional diffusion model. In: ICCV (2023)

\bibitem{zhang2024the}
Zhang, H., Zhou, J., Lu, Y., Guo, M., Wang, P., Shen, L., Qu, Q.: The emergence of reproducibility and consistency in diffusion models. In: ICML (2024)

\bibitem{zhang2022deis}
Zhang, Q., Chen, Y.: Fast sampling of diffusion models with exponential integrator. In: ICLR (2023)

\bibitem{zhao2023unipc}
Zhao, W., Bai, L., Rao, Y., Zhou, J., Lu, J.: Unipc: A unified predictor-corrector framework for fast sampling of diffusion models. In: NeurIPS (2023)

\bibitem{zhou2023fastamed}
Zhou, Z., Chen, D., Wang, C., Chen, C.: Fast ode-based sampling for diffusion models in around 5 steps. In: CVPR (2024)

\end{thebibliography}


\begin{thebibliography}{10}
\providecommand{\url}[1]{\texttt{#1}}
\providecommand{\urlprefix}{URL }
\providecommand{\doi}[1]{https://doi.org/#1}

\bibitem{binkowski2018demystifying}
Bi{\'n}kowski, M., Sutherland, D.J., Arbel, M., Gretton, A.: Demystifying mmd gans. In: ICLR (2018)

\bibitem{PFL}
Chen, C.: {PyTorch Face Landmark}: A fast and accurate facial landmark detector (2021)

\bibitem{dao2023flow}
Dao, Q., Phung, H., Nguyen, B., Tran, A.: Flow matching in latent space. arXiv preprint arXiv:2307.08698  (2023)

\bibitem{deng2019arcface}
Deng, J., Guo, J., Xue, N., Zafeiriou, S.: Arcface: Additive angular margin loss for deep face recognition. In: CVPR. pp. 4690--4699 (2019)

\bibitem{gu2020image}
Gu, J., Shen, Y., Zhou, B.: Image processing using multi-code gan prior. In: CVPR. pp. 3012--3021 (2020)

\bibitem{heusel2017gans}
Heusel, M., Ramsauer, H., Unterthiner, T., Nessler, B., Hochreiter, S.: Gans trained by a two time-scale update rule converge to a local nash equilibrium. In: NIPS. vol.~30 (2017)

\bibitem{Karras2022edm}
Karras, T., Aittala, M., Aila, T., Laine, S.: Elucidating the design space of diffusion-based generative models. NeurIPS  (2022)

\bibitem{karras2019style}
Karras, T., Laine, S., Aila, T.: A style-based generator architecture for generative adversarial networks. In: CVPR. pp. 4401--4410 (2019)

\bibitem{kastryulin2022piq}
Kastryulin, S., Zakirov, J., Prokopenko, D., Dylov, D.V.: Pytorch image quality: Metrics for image quality assessment (2022). \doi{10.48550/ARXIV.2208.14818}, \url{https://arxiv.org/abs/2208.14818}

\bibitem{kawar2022denoising}
Kawar, B., Elad, M., Ermon, S., Song, J.: Denoising diffusion restoration models. NeurIPS  \textbf{35},  23593--23606 (2022)

\bibitem{kawar2021snips}
Kawar, B., Vaksman, G., Elad, M.: Snips: Solving noisy inverse problems stochastically. In: NeurIPS. vol.~34, pp. 21757--21769 (2021)

\bibitem{kynkaanniemi2019improvedpr}
Kynk{\"a}{\"a}nniemi, T., Karras, T., Laine, S., Lehtinen, J., Aila, T.: Improved precision and recall metric for assessing generative models. In: NeurIPS. vol.~32 (2019)

\bibitem{lee2023minimizing}
Lee, S., Kim, B., Ye, J.C.: Minimizing trajectory curvature of ode-based generative models. In: ICML (2023)

\bibitem{lipman2022flow}
Lipman, Y., Chen, R.T., Ben-Hamu, H., Nickel, M., Le, M.: Flow matching for generative modeling. In: ICLR (2023)

\bibitem{liu2022flow}
Liu, X., Gong, C., et~al.: Flow straight and fast: Learning to generate and transfer data with rectified flow. In: ICLR (2023)

\bibitem{luhman2021knowledge}
Luhman, E., Luhman, T.: Knowledge distillation in iterative generative models for improved sampling speed. arXiv preprint arXiv:2101.02388  (2021)

\bibitem{obukhov2020torchfidelity}
Obukhov, A., Seitzer, M., Wu, P.W., Zhydenko, S., Kyl, J., Lin, E.Y.J.: High-fidelity performance metrics for generative models in pytorch (2020). \doi{10.5281/zenodo.4957738}, version: 0.3.0, DOI: 10.5281/zenodo.4957738

\bibitem{Peebles2022DiT}
Peebles, W., Xie, S.: Scalable diffusion models with transformers. In: ICCV (2023)

\bibitem{rombach2022high}
Rombach, R., Blattmann, A., Lorenz, D., Esser, P., Ommer, B.: High-resolution image synthesis with latent diffusion models. In: CVPR. pp. 10684--10695 (2022)

\bibitem{salimans2016improved}
Salimans, T., Goodfellow, I., Zaremba, W., Cheung, V., Radford, A., Chen, X.: Improved techniques for training gans. In: NIPS. vol.~29 (2016)

\bibitem{salimans2022progressive}
Salimans, T., Ho, J.: Progressive distillation for fast sampling of diffusion models. In: ICLR (2022)

\bibitem{song2023consistency}
Song, Y., Dhariwal, P., Chen, M., Sutskever, I.: Consistency models. In: ICML (2023)

\bibitem{song2021score}
Song, Y., Sohl-Dickstein, J., Kingma, D.P., Kumar, A., Ermon, S., Poole, B.: Score-based generative modeling through stochastic differential equations. In: ICLR (2021)

\bibitem{yu2023freedom}
Yu, J., Wang, Y., Zhao, C., Ghanem, B., Zhang, J.: Freedom: Training-free energy-guided conditional diffusion model. In: ICCV (2023)

\bibitem{zhang2024the}
Zhang, H., Zhou, J., Lu, Y., Guo, M., Wang, P., Shen, L., Qu, Q.: The emergence of reproducibility and consistency in diffusion models. In: ICML (2024)

\end{thebibliography}
% \end{raggedright}
\endgroup
\end{document}

\title{\textmd{Supplementary Material of Straighter Flow Matching via a Diffusion-Based Coupling Prior} }

\author{}
\institute{}
%
% \vspace{-2cm} 
%
\maketitle

This supplementary material  includes  experimental details, additional results and discussion.
We explain our experimental settings in Supp.~1.
We extend the application of {StraightFM} to the image inpainting task on FFHQ $256\times 256$ dataset in Supp.~2, thereby proving the efficacy of optimal transport coupling strategy for flow matching.
We conduct the ablation study in Supp.~3.
In Supp.~4, we summarize what makes StaightFM distinct from existing methods in the field of foundational generative models based on the experimental results.
Extensive results for image generation, multimodal conditional generation, and image inpainting are displayed in Supp.~5.

\section{Additional Implementation Details}
\subsection{ Implementation Details on the Pixel Space}
\subsubsection{Configuration}
 In configuring the velocity model $u_{\theta}$ and other hyperparameters, we fully reference the experimental setup used in 1-RectifiedFlow~\cite{liu2022flow} for CIFAR-10.
 Our implementation of StraightFM on CIFAR-10 
 is adapted from the open-source codes of ~\cite{song2021score,Karras2022edm,liu2022flow,lee2023minimizing}.
 For the architecture of $q_{\phi}(\tilde{\mathbf{x}}_0|\mathbf{x}_1)$ in Fast-ode+Ours, we adopt the configuration same as Fast-ode~\cite{lee2023minimizing}, where $q_{\phi}(\tilde{\mathbf{x}}_0|\mathbf{x}_1)$ is also a smaller UNet than the velocity model.
The pseudo-image and noise couplings are constructed online.
  We test two ODE solvers for StraightFM sampling process: the Euler method for fixed-step sampling and the Runge-Kutta method for adaptive-step sampling.
Other training configuration is comprehensively outlined in \cref{tab:config}.
 The quality and diversity of results are evaluated according to IS (Inception Score~\cite{salimans2016improved}) and FID (Fr\'echet Inception Distance~\cite{heusel2017gans}),  following the code of ScoreSDE~\cite{song2021score}.
 % ~ 
 \begin{table*}[!bthp]
\centering
\caption{Configuration on CIFAR-10, CelebA-HQ and FFHQ.}
\label{tab:config}
\resizebox{\textwidth}{!}{%
\begin{tabular}{@{}c|cc|cc|c@{}}
\toprule
     & \multicolumn{2}{c|}{CIFAR-10}           & \multicolumn{2}{c|}{CelebA-HQ 256$\times$ 256} & FFHQ 256$\times$ 256 \\
                    & StraighFM & Fast-ode+Ours & StraighFM &  Fast-ode+Ours & Inpainting Task \\ \midrule
Architecture        & NCSN++      & NCSN++       & DiT-L/2         & DiT-L/2          & NCSN++          \\
Learning rate       & 2e-4        & 2e-4         & 5e-4        & 1e-5         & 2e-5            \\
Batch size          & 4$\times$64 & 5$\times$64  & 4$\times$4 & 4$\times$4  & 2$\times$16     \\
$\lambda$           & 0           & 10           & 0           & 10           & 0               \\
Training iterations & 600K        & 700K         & 800       & 650         & 20K             \\
EMA decay rate      & 0.999999    & 0.999999     & 0.9999      & 0.9999       & 0.999           \\
GPUs & 4$\times$ RTX 3090 & 5$\times$ RTX 4090 & 4$\times$Tesla V100    & 4$\times$Tesla V100   & 2$\times$TITAN RTX   \\
Dropout probability & 0.15        & 0.15         & 0.0         & 0.0          & 0.0             \\ \bottomrule
\end{tabular}%
}
\end{table*}

\subsubsection{{Cost Comparison}}
We display the training cost, one-step generation ability, and inference speed of StraightFM and existing open-source generative models in~\cref{tab:cost}.
StraightFM, with the one-step generation ability, is achieved by fewer iterations and consumer-level GPUs compared to alternative methods. 
Our approach demonstrates improved efficiency by requiring fewer iterations than other competitive methods.

\begin{table}[!bhtp]
\centering
\caption{Comparison of the training cost and inference speed of different generative models on CIFAR-10.}
 \renewcommand{\arraystretch}{0.6}
{%
\resizebox{\textwidth}{!}{
\begin{tabular}{@{}lllcccc@{}}
\toprule
     Method & GPUs           & \begin{tabular}[c]{@{}l@{}} Training iterations\end{tabular}                         & \begin{tabular}[c]{@{}l@{}}One step\end{tabular} & \begin{tabular}[c]{@{}l@{}} Speed (per step)\end{tabular} &Network structure
     \\ \midrule
Consistency Model~\cite{song2023consistency}   & 8$\times$ A100 & 800K       & $\checkmark$        & 0.033s   &NCSN++        \\
1-Rectified Flow~\cite{liu2022flow} & 8$\times$ A100 & 800K                               & $\times$        & 0.031s &DDPM++           \\
2-Rectified Flow~\cite{liu2022flow} & 8$\times$ A100 & 300K+800K                           & $\checkmark$        & 0.030s &DDPM++           \\
Ours & 4$\times$ RTX 3090    & 600K                                 & $\checkmark$  & {0.030s}&DDPM++      \\ \bottomrule
\end{tabular}%
}}\label{tab:cost}
\end{table}

\subsection{Implementation Details in the Latent Space}% 
\subsubsection{Configuration}
Our implementation in the latent space of Latent Diffusion Model on CelebA-HQ $256\times 256$ 
builds upon the open-source of LFM (Latent Flow Matching)~\cite{dao2023flow}.
We offline generated about 280K pseudo-image and noise couplings to construct the training dataset.
The architecture of $q_{\phi}(\tilde{\mathbf{x}}_0|\mathbf{x}_1)$ follows Fast-ode~\cite{lee2023minimizing}.
The structure of the velocity model utilizes DiT~\cite{Peebles2022DiT}.
The parameters are updated using the exponential moving average (EMA) with a decay rate of 0.999. 
Other training configuration is  outlined in \cref{tab:config}.

 Results are evaluated according FID (Fr\'echet Inception Distance~\cite{heusel2017gans}) and KID (Kernel Inception Distance~\cite{binkowski2018demystifying}), utilizing an open-source assessment torch-fidelity~\cite{obukhov2020torchfidelity}.
 KID compares the squared Maximum Mean Discrepancy with polynomial kernel between the distribution of real and generated images by comparing the features extracted using the same method as FID.
The lower FID, the better generative capability.
 We totally generate 10,000 samples from each method to calculate FID and KID metrics.
\subsubsection{Baselines}
 StraightFM is mainly compared with Rectified Flow (RF)~\cite{liu2022flow}, Latent Diffusion Model (LDM)~\cite{rombach2022high}, and Latent Flow Matching (LFM, training flow matching in the latent space)~\cite{dao2023flow} on CelebA-HQ dataset. 
%  % 
 Rectified Flow is trained on the pixel space following the structure of ScoreSDE~\cite{song2021score}, while LFM and LDM are trained on the latent space of the pre-trained  autoencoder models~\cite{rombach2022high}.
LDM sampling is performed with a DDIM sampler, while StraightFM and other baselines utilize  Euler method with a uniform time-step scheme.

\subsection{Implementation Details on Conditional Generation}\label{supp:sub-freedom}
\hspace{0.5em}
We  employ FreeDoM as the baseline method.
FreeDoM is an open-source implementation of a general conditional guided sampling framework specially designed for diffusion model~\cite{yu2023freedom}.
We evaluate StraightFM to text-guided, landmark-guided, and facial ID-guided generation.
According to Eq.(8), the off-the-shelf pre-trained models for various condition are same as the default setting of FreeDoM:
For text-guided sampling, we use the $\ell_2$ Euclidean distance function based on CLIP text encoder and CLIP image encoder.
An open-source pre-trained landmark detection network~\cite{PFL} is used  for landmark-guided sampling.
We utilize an open-source pre-trained face recognition network to generate a identity  to extract representation from reference images and generated images~\cite{deng2019arcface}.

\section{ Additional Experiment: Inpainting Task}\label{supp:sub-inp}% 

Flow matching methods have been applied in various tasks of image processing, including super-resolution~\cite{lipman2022flow} and domain transfer, e.g., from cats to dogs~\cite{liu2022flow}.
For image restoration tasks, the pairs of degraded and target images naturally form optimal transport couplings, making StraightFM equivalent to naive flow matching, where the source distribution is not constrained in Gaussian distribution.
As a complement to this study, we validate the effectiveness of flow matching in image restoration task, particularly in the inpainting task, addressing the missing parts of images on the FFHQ 256$\times$256~\cite{karras2019style} dataset.

\subsubsection{Implementation Details}
Considering that these corresponding image pairs are natural optimal transport couplings, it is feasible to directly train a flow matching model using them.
We conduct the inpainting task on FFHQ $256\times 256$ dataset. 
In this procedure, we randomly mask 50\%  of the pixel values in the original images to create degraded images.
After that, these degraded images serve as source samples from $p_0$, while corresponding original images can be regarded as target samples from $p_1$.
Results are measured by PSNR (Peak Signal-to-Noise Ratio), SSIM (Structural Similarty), and LPIPS (Learned Perceptual Image Patch Similarity). 

For dataset splitting, FFHQ dataset, which consists of 70,000 high-resolution, well-aligned face images, is partitioned into two segments:  the first 60,000 images for training and the remaining 10,000 for validation.
 The image quality assessment in this inpainting task is conducted with PyTorch Image Quality~\cite{kastryulin2022piq}.

\subsubsection{Baselines}
Under the same degradation configuration, our competitors include SNIPS and mGANPrior.
SNIPS~\cite{kawar2021snips} is not originally designed for FFHQ dataset. 
The initial work in SNIPS directly calculates Singular Value Decomposition (SVD), which hinders its ability to process images with a resolution of 256$\times$ 256. 
To facilitate comparison, we replace this direct calculation with an efficient implementation, leveraging notable properties of SVD to reduce the space complexity to $O(n)$~\cite{kawar2022denoising}.

Another competitor, mGANPrior~\cite{gu2020image}, employs a technique of optimizing multiple latent codes at certain intermediate layers of a pre-trained generator to generate feature maps for recovering the input image.
Our reimplementation of mGANPrior is based on StyleGAN~\cite{karras2019style}, which is trained on the entire FFHQ dataset.
 Further configuration details are available in \cref{supp:tab:ffhq_inp}.
 
\begin{table}[!htbp]
\centering
\caption{Hyper-parameters of mGANPrior on inpainting task.}
\label{supp:tab:ffhq_inp}
\begin{tabular}{@{}cc@{}}
\toprule
     Configuration              & mGANPrior       \\ \midrule
Inversion type     & StyleGAN-w+ \\
Composing layer    & 4           \\
Latent code number & 30          \\
Iterations         & 3000        \\
Loss  type             & L2+VGG      \\
Optimization       & Adam        \\ \bottomrule
\end{tabular}
\end{table}

   \begin{figure*}[htp]
\centering
\begin{minipage}{0.48\textwidth}
	\begin{subfloat}
			\centering \includegraphics[scale=0.15]{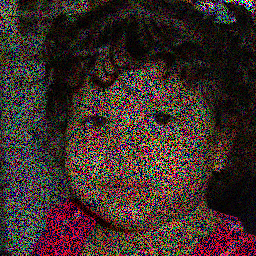}
		\end{subfloat}
  \hspace{-2mm}
	\begin{subfloat}
			\centering \includegraphics[scale=0.15]{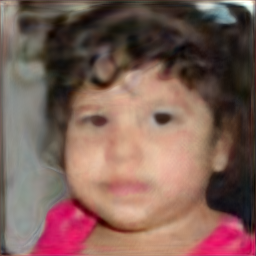}
		\end{subfloat}    \hspace{-2mm}
	\begin{subfloat}
			\centering \includegraphics[scale=0.15]{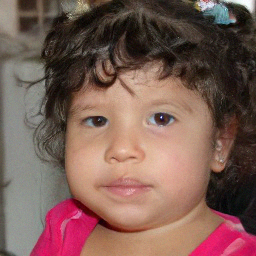}
		\end{subfloat}   \hspace{-2mm}
	\begin{subfloat}
			\centering \includegraphics[scale=0.15]{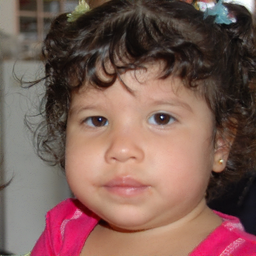}
		\end{subfloat}

	\begin{subfloat}
			\centering \includegraphics[scale=0.15]{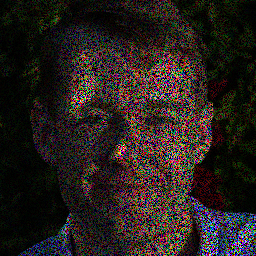}
		\end{subfloat}
 \hspace{-2mm}
	\begin{subfloat}
			\centering \includegraphics[scale=0.15]{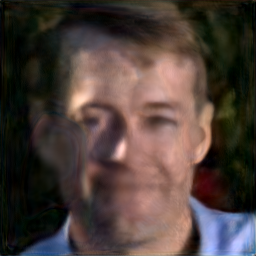}
		\end{subfloat}  \hspace{-2mm}
  	\begin{subfloat}
			\centering \includegraphics[scale=0.15]{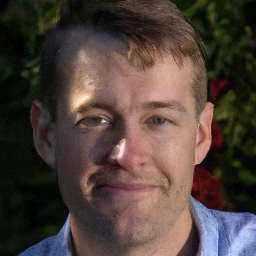}
		\end{subfloat}  \hspace{-2mm}
  	\begin{subfloat}
			\centering \includegraphics[scale=0.15]{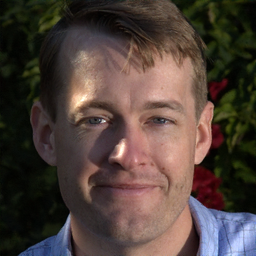}
		\end{subfloat}\vspace{-1.0mm}

\hspace{2mm}\begin{subfloat}\centering 
		\hspace{0mm}{\small	Input }
		\end{subfloat}
	\begin{subfloat}\centering  	\hspace{0mm}{\small
mGANPrior}
  \end{subfloat}
  	\begin{subfloat}\centering {\small
SNIPS}		\end{subfloat} 
  	\begin{subfloat} 
			\centering{\small \hspace{1mm} Ours}
		\end{subfloat}
		\vspace{-2.5mm}
  \caption{Image inpainting results on  FFHQ $256\times 256$ dataset. }		  \label{fig:ffhq-inp}
\end{minipage}    \hfill
\begin{minipage}{0.48\textwidth}
    \centering
    \captionof{table}{Quantitative comparison of inpainting results. $\uparrow$ means higher is better, and $\downarrow$ means lower is better. }
  \renewcommand{\arraystretch}{0.44}
      \resizebox{1\textwidth}{!}{
\begin{tabular}{@{}lrccc@{}}
\toprule
          & PSNR$\uparrow$ & SSIM$\uparrow$ & LPIPS $\downarrow$ \\ \midrule
mGANPrior & 23.82              & 0.71              & 0.43                                          \\
SNIPS     & 30.30       & 0.65          & 0.12                                    \\
Ours      & \textbf{37.43}         & \textbf{0.97}             & \textbf{0.06}                                     \\ \bottomrule
\end{tabular}%
}
\label{tab:ffhq-inp}
\end{minipage}
\centering
\end{figure*}

 \subsubsection{Results}
\Cref{fig:ffhq-inp} shows the restored images of StraightFM in the inpainting task, compared with mGANPrior~\cite{gu2020image} and SNIPS~\cite{kawar2021snips}, whose degradation configurations are the same as ours.
Quantitative results are shown in \cref{tab:ffhq-inp}, which demonstrates the potential of an effective coupling strategy, supporting our argument about the significance of coupling strategy in flow matching performance.

 \section{Ablation Study}\label{subsec:ablation}
  We conduct an ablation study to discern whether the effective sampling performance of StraightFM is due to the diffusion-based coupling prior.
 We aim to analyze the two factors related to StraightFM: 1)
 coupling strategy, 2) the data scale of couplings provided by diffusion prior.
We also declare that we do not consider the impact of diffusion model size in our work because a previous study~\cite{song2021score}  demonstrated that varying the network structure could not bring significant improvement.

 \subsubsection{{Coupling Strategy}}
To compare with StraightFM, we implement flow matching using exclusively couplings from real data to noise with two different strategies: forward diffusion (the forward diffusion process~\cite{song2021score}) and parameterization with neural network (Fast-ode).

 \begin{table}[!thbp]
 \caption{The performance of different coupling strategies on CIFAR-10. We reproduce the result of Neural net~\cite{lee2023minimizing}. $\uparrow$ means higher is better, and $\downarrow$ means lower is better. }
\label{tab:tab2}
\centering
\renewcommand{\arraystretch}{0.65}
{%
\begin{tabular}{@{}lcc@{}}
\toprule
Coupling strategy   & FID$\downarrow$ / IS$\uparrow$ (N=1) & FID$\downarrow$ / IS$\uparrow$ (N=adap.) \\ \midrule
Forward diffusion &    375/1.14        & 3.97/9.18              \\
Neural net.   & 240/2.10     & 3.83/9.25        \\
Neural net.+Ours   & {\ul 29.4}/{\ul 7.41}     & \textbf{2.82}/\textbf{9.52}        \\
Ours  & \textbf{8.65}/\textbf{8.56}     &{\ul 2.94}/{\ul 9.51}        \\
\bottomrule
\end{tabular}%
} 
\end{table}

\Cref{tab:tab2} presents how the coupling strategy affects the sample quality. 
The coupling strategy of StraightFM is the key factor to few-step generation, significantly improving the image generation performance in terms of IS and FID scores.
The combination of StraightFM and the parametrization of forward  diffusion direction with a neural network allows for better image quality in one-step sampling than the original single forward diffusion direction.
Additional ablation study is embedded within some experimental components above. 
Specifically, competing methods, e.g., RF, FM, I-CFM on CIFAR-10 in subsection~5.1 and  LFM on CelebA-HQ in subsection~5.2, take a totally random coupling strategy to train flow matching models.

\begin{figure*}[!thbp]
    \hspace{3mm}
     \includegraphics[scale=0.38]{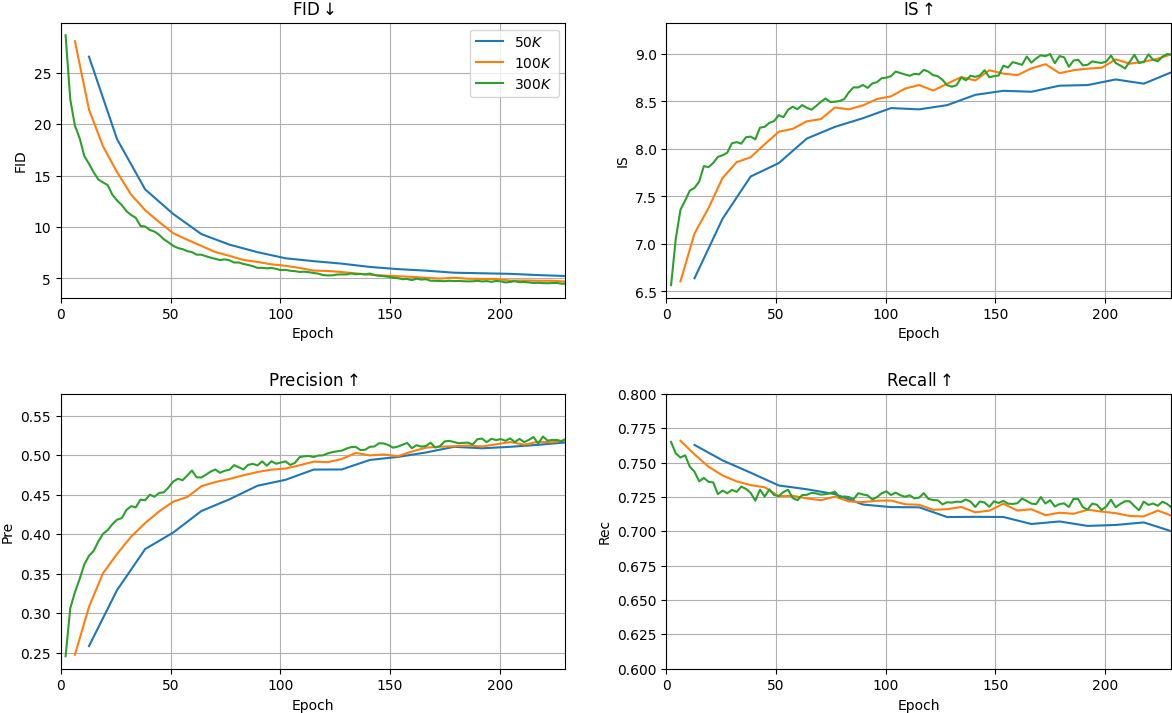}
     \caption{Relationship between the quality of generated samples (FID, IS, Precision, and Recall) and the scale of diffusion prior at same training epoch. The sampling step is set as 3. $\uparrow$ means higher is better, and $\downarrow$ means lower is better.  }
     \label{fig:ablationscale}
 \end{figure*}
 \subsubsection{{Data Scale}}
The goal of generative methods is to learn the manifold of the real data so that we can subsequently
generate novel samples that are indistinguishable from real samples.
The training set of StraightFM consists of diffusion couplings. 
To explore whether StraightFM captures the real data distribution when varying the data scale, we set the diffusion coupling sizes to 50K, 100K, and 300K, respectively.
The evaluation metrics include FID and IS, as well as Precision and Recall~\cite{kynkaanniemi2019improvedpr}, which individually assess the quality and coverage of the generated samples. 
A higher value of Precision indicates that the generated images are closer to real data manifold, and a higher value of Recall implies that the generated images can capture  broader diversity present in real data.

A clear pattern from~\cref{fig:ablationscale} is: when the data scale is larger, the performance of StraightFM may initially seem like getting an improvement.
 However, when the epoch increases, the FID value at different scales are almost indistinguishable, as well as the Precision value.
At around 200 epochs, the variation of IS  at different data scales is less than 0.4, and the  Recall values at different data scales fluctuate within the range of approximately  between 0.7 and 0.725.
It is reasonable as just sufficient couplings could help StraightFM learn real data distribution. 
As a result, a larger diffusion prior would not obviously improve sample quality.
 The result suggests that the generalization ability of StraightFM is thanks to the unique coupling strategy in training,  and it does not rely on a large enough data scale that is analogous to image augmentation. 

\section{Additional Discussion}\label{sec:discuss}
Based on the experimental results, we summarize what makes StaightFM distinct from existing methods in the field of foundational generative models.

Distillation technique is a predominant direction to speed up the
sampling speed of existing diffusion models.
Several works distillate a large quantity of sampling trajectories at different timesteps from diffusion model to obtain a fast one~\cite{luhman2021knowledge,salimans2022progressive,song2023consistency}.
Although the technique seems to resemble StraightFM, there are several key differences from our proposed method:
% 训练上面
\begin{itemize}[leftmargin=*]
    \item \textit{Training procedure}: StraightFM only needs samples at the initial and final timesteps  from diffusion prior to train a flow matching model, which obviously distinguishes it from existing distillation based methods.
The comparison of the training cost (in \cref{tab:cost}) and training set scale (in \cref{fig:ablationscale}) also demonstrates  that our approach can effectively learn real data distribution, rather than   relying on a larger training dataset and computational resource.
\item \textit{Generalization ability}:
Distillation-based methods usually obtain a student model as a one-step generator, disturbing the inherent structure in the original diffusion model. 
This student model  with huge training cost usually can not easily generalize to  some training-free application such as conditional guided sampling~\cite{yu2023freedom}, i.e., consistency model~\cite{song2023consistency}.
Our approach inherits from flow matching, keeping the formulation of ODE. As an unconditional model,   StraightFM is seamlessly compatible with training-free multimodal conditional generation that is the same as diffusion models. 
\item \textit{The necessity of flow matching}:  Our motivation is based on the special property of the coupling similarity between independent  diffusion  model and flow matching methods, which is absent in other foundational generative models like GAN and VAE~\cite{zhang2024the}.
Due to this property, StraightFM bypasses the need for solving OT problems in minibatches or multi-round training.
Flow matching is crucial and necessary to stand out, as it explicitly constructs the straight trajectories. 
Otherwise, it risks being overshadowed by the hundreds of iterations nature 
and the associated computational costs of diffusion model.
\end{itemize}

\section{More results}
\begin{figure*}
    \centering
    \includegraphics[scale=0.6]{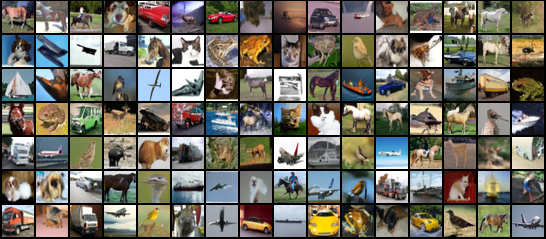}
    \caption{Uncurated samples from StraightFM on CIFAR-10.}
    \label{fig:cifar10-1}
\end{figure*}

\begin{figure*}
    \centering
    \includegraphics[scale=0.6]{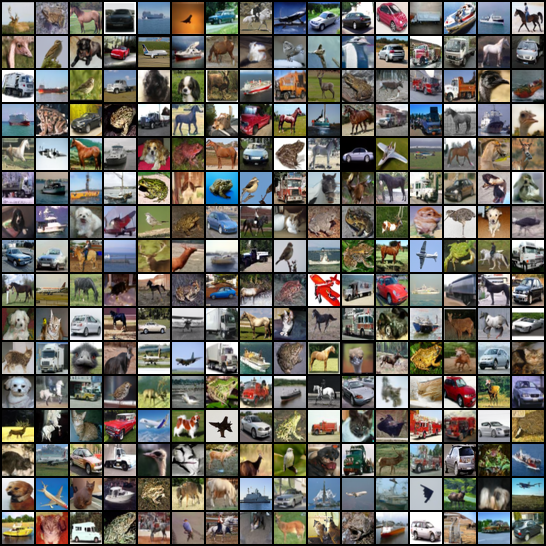}
    \caption{Uncurated samples from Fast-ode+Ours on CIFAR-10.}
    \label{fig:cifar10-2}
\end{figure*}

\begin{figure*}[!t]
    \centering
    \includegraphics[scale=0.50]{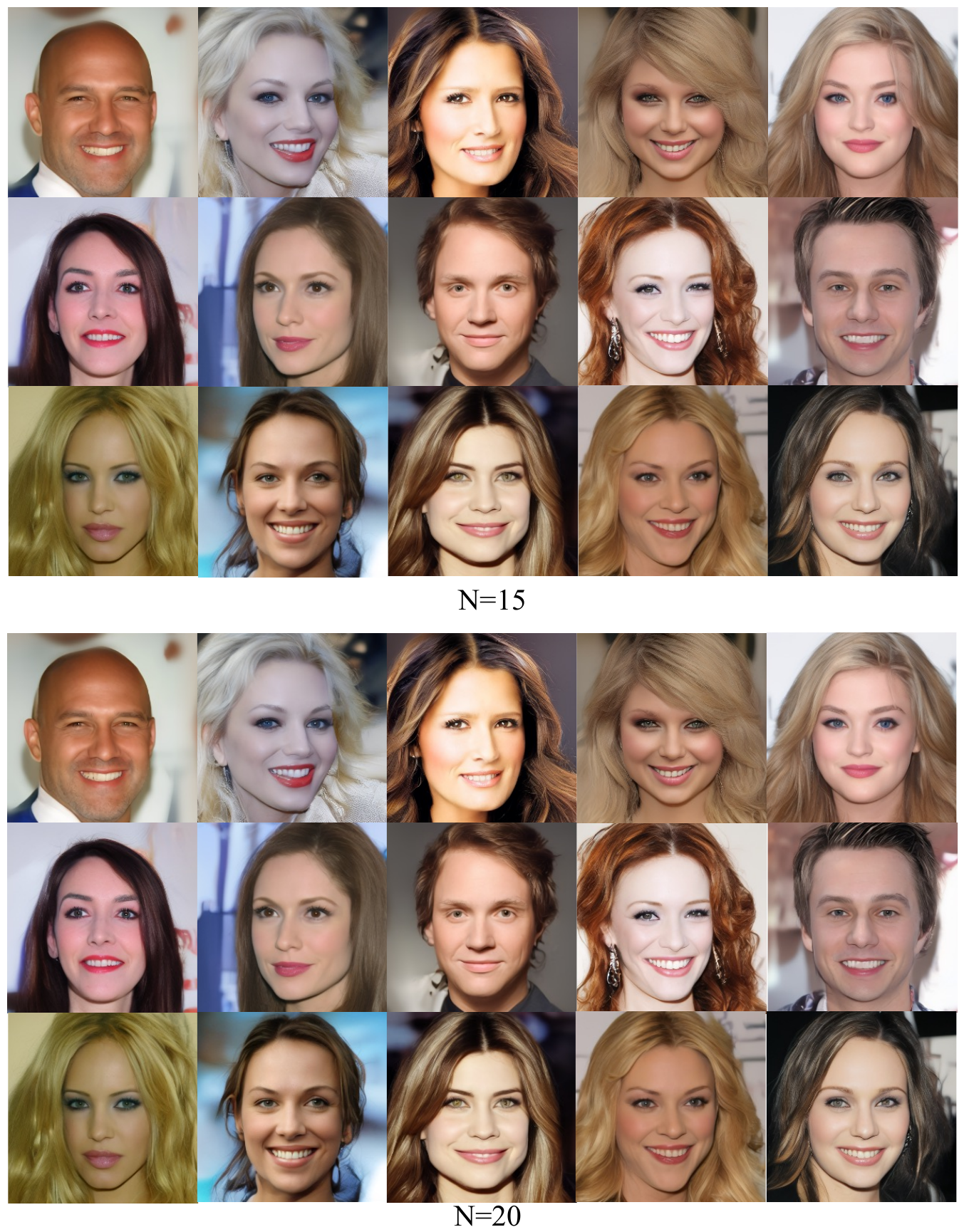}
    \caption{Samples from StraightFM on CelebA-HQ.}
    \label{suppfig:celebahq-col6}
\end{figure*}

\begin{figure*}[!t]
    \centering
    \includegraphics[scale=0.50]{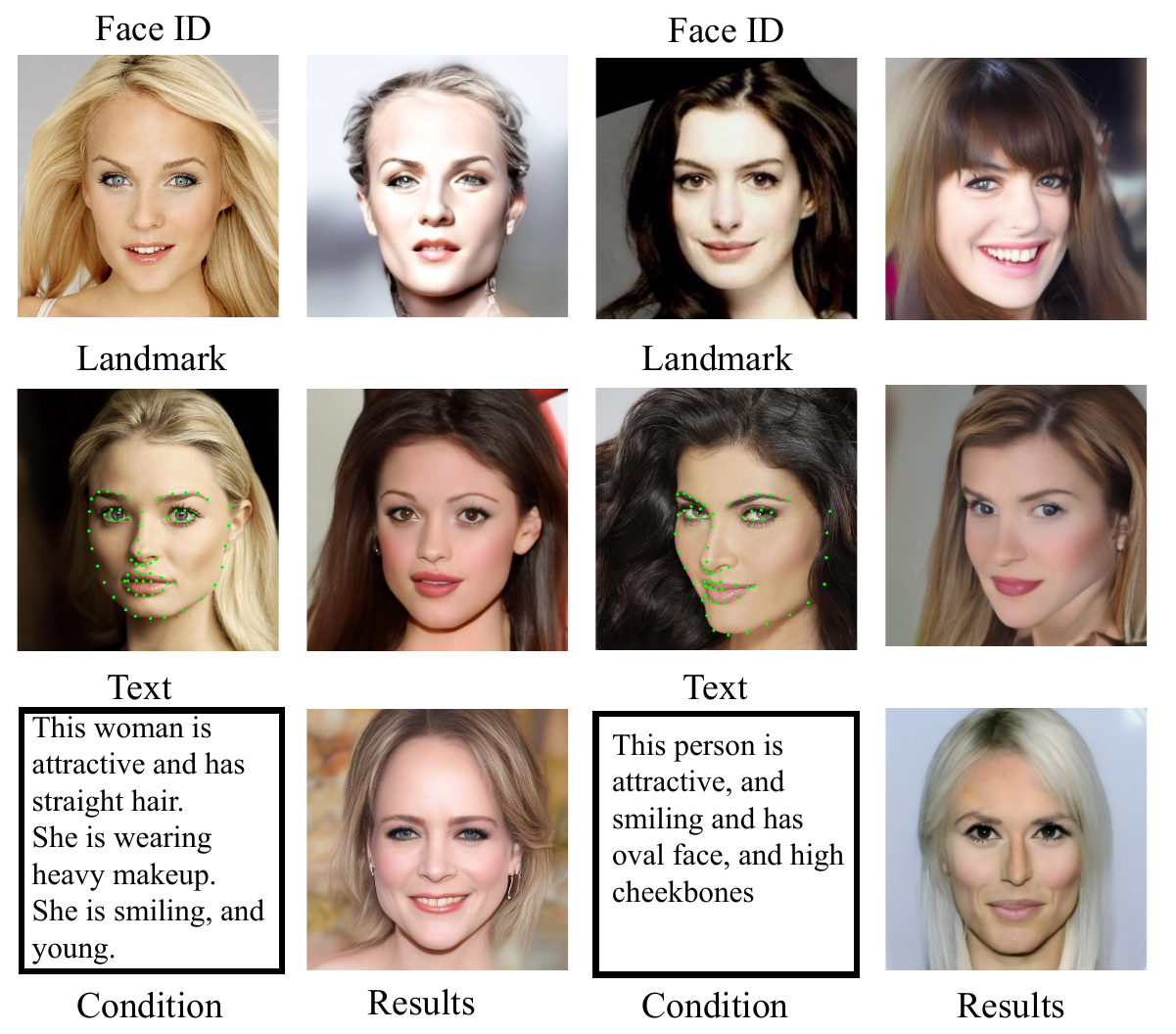}
    \caption{Multimodal conditional generation with guided sampling.}
    \label{suppfig:cond_celebahq}
\end{figure*}

\begin{figure*}[!t]
\centering
	\begin{subfloat}
			\centering	\includegraphics[scale=0.3]{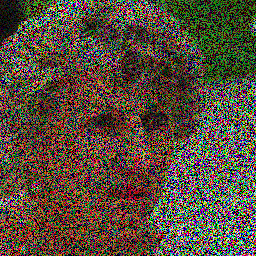}
\end{subfloat}
	\begin{subfloat}
			\centering \includegraphics[scale=0.3]{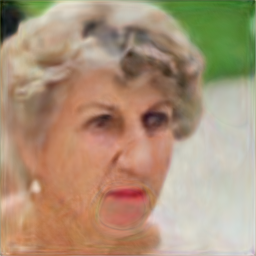}
		\end{subfloat}
	\begin{subfloat}
			\centering \includegraphics[scale=0.3]{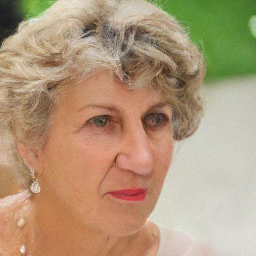}
		\end{subfloat}
	\begin{subfloat}
			\centering \includegraphics[scale=0.3]{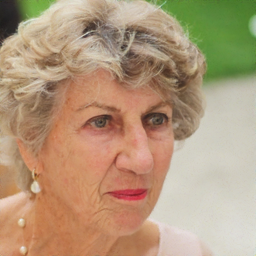}
		\end{subfloat}\vspace{1mm}

	\begin{subfloat}
			\centering	\includegraphics[scale=0.3]{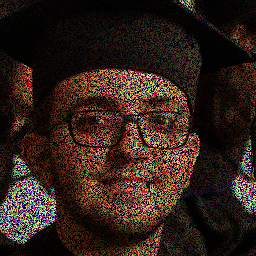}
\end{subfloat}
	\begin{subfloat}
			\centering \includegraphics[scale=0.3]{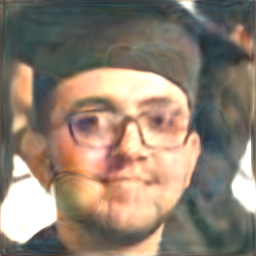}
		\end{subfloat}
	\begin{subfloat}
			\centering \includegraphics[scale=0.3]{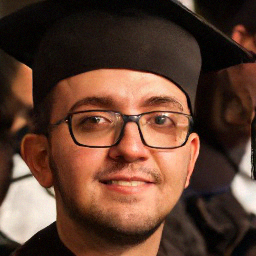}
		\end{subfloat}
	\begin{subfloat}
			\centering \includegraphics[scale=0.3]{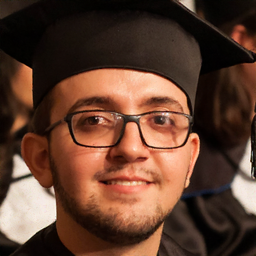}
		\end{subfloat}\vspace{1mm}

	\begin{subfloat}
			\centering	\includegraphics[scale=0.3]{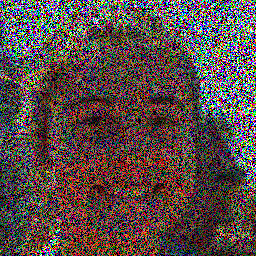}
\end{subfloat}
	\begin{subfloat}
			\centering \includegraphics[scale=0.3]{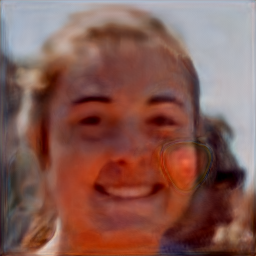}
		\end{subfloat}
	\begin{subfloat}
			\centering \includegraphics[scale=0.3]{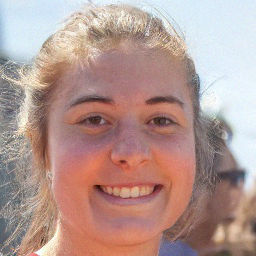}
		\end{subfloat}
	\begin{subfloat}
			\centering \includegraphics[scale=0.3]{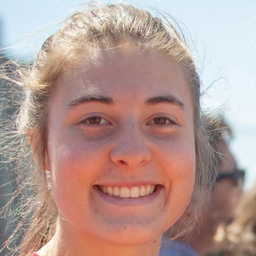}
		\end{subfloat}\vspace{1mm}

	\begin{subfloat}
			\centering	\includegraphics[scale=0.3]{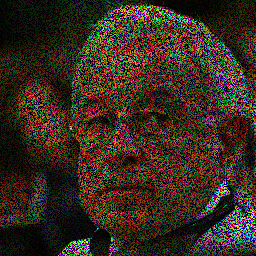}
\end{subfloat}
	\begin{subfloat}
			\centering \includegraphics[scale=0.3]{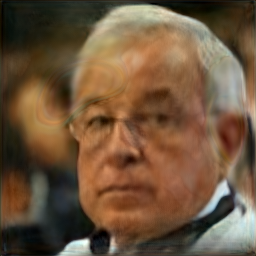}
		\end{subfloat}
	\begin{subfloat}
			\centering \includegraphics[scale=0.3]{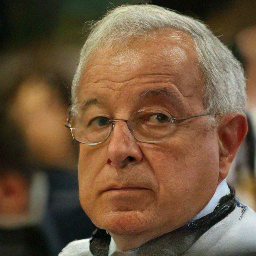}
		\end{subfloat}
	\begin{subfloat}
			\centering \includegraphics[scale=0.3]{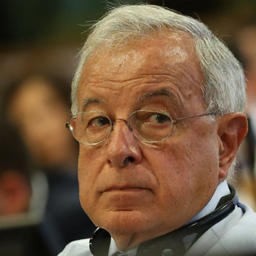}
		\end{subfloat}\vspace{1mm}

  	\begin{subfloat}
			\centering	\includegraphics[scale=0.3]{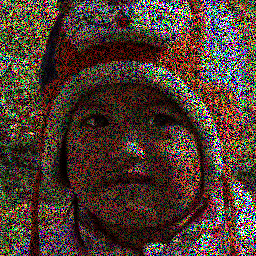}
\end{subfloat}
	\begin{subfloat}
			\centering \includegraphics[scale=0.3]{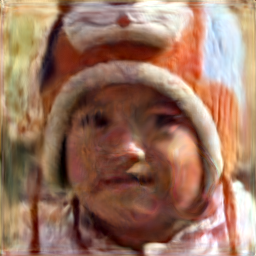}
		\end{subfloat}
	\begin{subfloat}
			\centering \includegraphics[scale=0.3]{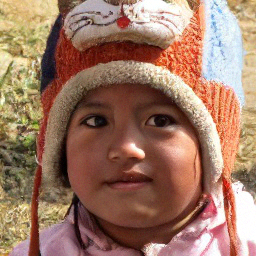}
		\end{subfloat}
	\begin{subfloat}
			\centering \includegraphics[scale=0.3]{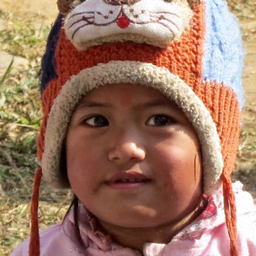}
		\end{subfloat}\vspace{1mm}

  	\begin{subfloat}
			\centering	\includegraphics[scale=0.3]{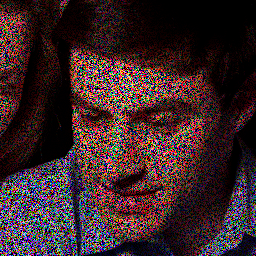}
\end{subfloat}
	\begin{subfloat}
			\centering \includegraphics[scale=0.3]{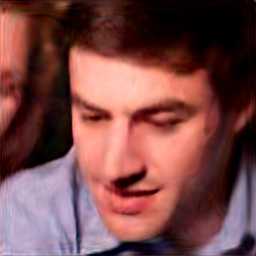}
		\end{subfloat}
	\begin{subfloat}
			\centering \includegraphics[scale=0.3]{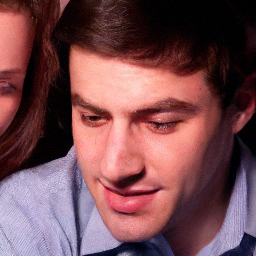}
		\end{subfloat}
	\begin{subfloat}
			\centering \includegraphics[scale=0.3]{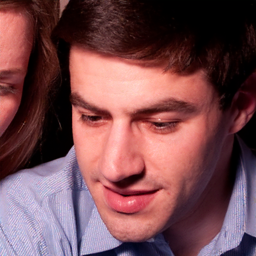}
		\end{subfloat}\vspace{1mm}

\hspace{0cm}
  	\begin{subfloat}
			\centering	degraded
\end{subfloat}
\hspace{1cm}
	\begin{subfloat}
			\centering  mGANPrior
		\end{subfloat}
  \hspace{1cm}
	\begin{subfloat}
			\centering SNIPS
		\end{subfloat}
  \hspace{2cm}
	\begin{subfloat}
			\centering Ours
		\end{subfloat}
%   \vspace{-3mm}
% \vspace{2mm}
    \caption{Uncurated samples from FFHQ on inpainting task.}
    \label{fig:suppffhq}
\end{figure*}

To better illustrate,  additional synthesized examples from  CIFAR-10 dataset are showcased in \cref{fig:cifar10-1} and \cref{fig:cifar10-2}.
 For  CelebA-HQ 256 dataset, please refer to \cref{suppfig:celebahq-col6}, and \cref{suppfig:cond_celebahq}, and for  FFHQ 256 dataset, relevant examples can be found in	  \cref{fig:suppffhq}.
 

\bibliographystyle{splncs04}
\bibliography{straightfm.bib}